\documentclass[acmtog]{acmart}
\acmSubmissionID{352}
\AtBeginDocument{%
  }

\setcopyright{acmlicensed}
\copyrightyear{2026}
\acmYear{2026}
\setcopyright{cc}
\setcctype{by}
\acmConference[SIGGRAPH Conference Papers '26]{Special Interest Group on Computer Graphics and Interactive Techniques Conference Conference Papers}{July 19--23, 2026}{Los Angeles, CA, USA}
\acmBooktitle{Special Interest Group on Computer Graphics and Interactive Techniques Conference Conference Papers (SIGGRAPH Conference Papers '26), July 19--23, 2026, Los Angeles, CA, USA}
\acmDOI{10.1145/3799902.3811066}
\acmISBN{979-8-4007-2554-8/2026/07}

\usepackage{xcolor}
\usepackage[normalem]{ulem}
\usepackage{multirow}
\usepackage{booktabs}
\usepackage{multirow}
\usepackage{makecell}
\usepackage{tcolorbox}

\newcommand{\z}[1]{{\color{magenta} #1}}

\newcommand{\warning}[1]{{\color{red} \sout {#1}}}

\begin{document}

\title{EchoAvatar: Real-time Generative Avatar Animation from Audio Streams}

\author{Bohong Chen}
\email{bohongchen@zju.edu.cn}
\orcid{0009-0007-1036-7737}
\affiliation{
\institution{State Key Lab of CAD\&CG, Zhejiang University}
\city{Hangzhou}
\country{China}
}

\author{Yumeng Li}
\email{yumeng.li@zju.edu.cn}
\orcid{0009-0007-6558-4165}
\affiliation{
\institution{State Key Lab of CAD\&CG, Zhejiang University}
\city{Hangzhou}
\country{China}
}

\author{Yinglin Xu}
\email{ylxu@zju.edu.cn}
\orcid{0009-0008-9804-3872}
\affiliation{
\institution{State Key Lab of CAD\&CG, Zhejiang University}
\city{Hangzhou}
\country{China}
}

\author{Youyi Zheng}
\email{youyizheng@zju.edu.cn}
\orcid{xxx}
\affiliation{
\institution{State Key Lab of CAD\&CG, Zhejiang University}
\city{Hangzhou}
\country{China}
}

\author{Yanlin Weng}
\email{weng@cad.zju.edu.cn}
\orcid{xxx}
\affiliation{
\institution{State Key Lab of CAD\&CG, Zhejiang University}
\city{Hangzhou}
\country{China}
}

\author{Kun Zhou}
\authornote{Corresponding author}
\email{kunzhou@acm.org}
\orcid{0000-0003-4243-6112}
\affiliation{
\institution{State Key Lab of CAD\&CG, Zhejiang University}
\city{Hangzhou}
\country{China}
}



\begin{teaserfigure}
  \centering
  \includegraphics[width=\textwidth]{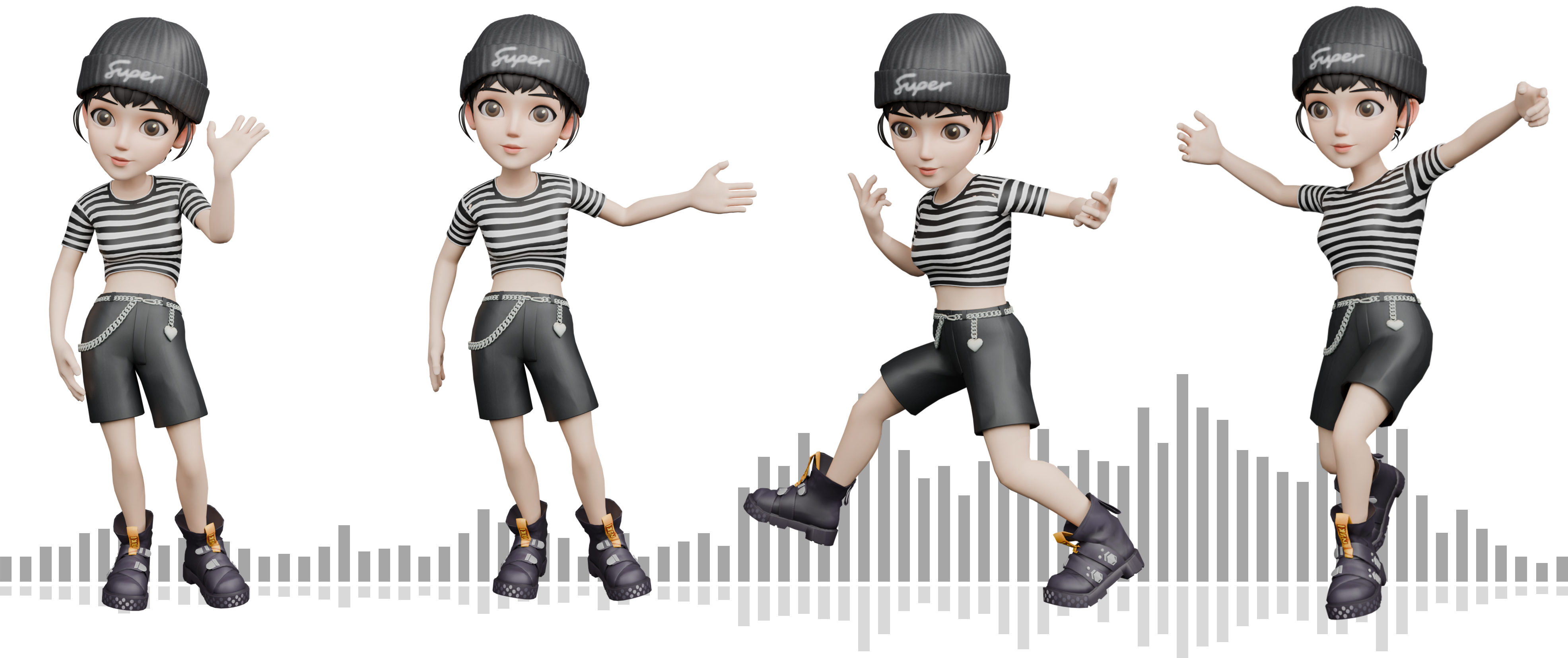}
  \caption{
 Given streaming audio input, our method generates avatar animation in a streaming manner. The four poses shown above are sampled from a continuous motion sequence driven by the audio stream.
  }
  \label{fig:teaser}
\end{teaserfigure}



\begin{abstract}

Real-time synthesis of high-fidelity 3D character motion from audio is a pivotal component for next-generation interactive avatars and virtual assistants. However, most existing approaches are limited to offline processing of complete audio sequences or are constrained to specific domains, rarely handling both speech and music effectively. {In this paper, we introduce a novel framework designed to generate continuous, coherent full-body motion from streaming speech and music with low latency}. Central to our approach is a unified streaming architecture capable of synthesizing continuous motion from incremental audio inputs. We employ a robust training strategy that enforces strong audio dependency, allowing the model to seamlessly generalize across conversational speech and rhythmic music without requiring explicit domain labels or mode switching. Additionally, we explored Reinforcement Learning to refine the quality of online generation. Furthermore, we bridge reactive animation with intent-driven behavior via a tool-call interface that allows upstream Large Language Models to inject explicit semantic control. By combining this controllability with stream audio-driven synthesis, our framework serves as a plug-and-play solution for transforming voice agents into interactive humanoid avatars. Extensive experiments demonstrate that our method outperforms state-of-the-art real-time baselines in motion quality and synchronization while maintaining the flexibility required for live deployment. Our code, pre-trained models, and videos are available at https://robinwitch.github.io/EchoAvatar-Page.
\end{abstract}

\ccsdesc[500]{Computing methodologies~Animation}
\ccsdesc[500]{Computing methodologies~Artificial intelligence}

\keywords{Streaming Motion Generation}

\maketitle

\section{Introduction}

The rapid evolution of Large Language Models (LLMs) and Voice Agents has enabled fluid, natural dialogue. However, while audio fidelity is near-human, visual embodiment lacks the responsiveness required for genuine interaction, necessitating high-fidelity 3D motion generation directly from streaming audio with low latency.

Existing approaches to audio-driven motion synthesis fall short of meeting this challenge for two primary reasons. First, most state-of-the-art methods~\cite{liu2024emage,Zhang2024SemanticGesture,chen2025meco} are designed for offline processing, requiring complete audio sequences as input before generating motion. This architectural constraint introduces unacceptable latency for real-time interactive applications. Second, existing methods are typically domain-specific, handling either speech or music but rarely both. This fragmentation necessitates complex model switching, limiting their applicability to general-purpose voice agents that must handle diverse acoustic inputs uniformly.

In this paper, we present a unified framework for real-time, streaming avatar animation that addresses these limitations. This system takes live audio stream and generates motion in real-time, featuring a causal motion tokenizer for high-quality auto-regressive synthesis and a specially designed training strategy for unified learning. Regarding the tokenizer, while recent works have explored causal architectures via causal convolutions~\cite{jiang2025causal,Xiao2025ICCV}, we find that pure convolutional approaches often lack expressiveness and suffer from reconstruction artifacts. We instead propose an attention-based causal motion tokenizer with auxiliary kinematic losses, achieving superior generation quality and streaming capability. Regarding the training strategy, we identify that optimization dynamics within a unified motion space often weaken audio conditioning, leading to catastrophic failure in task alignment. We address this via a hierarchical token corruption strategy that enhances audio conditioning, enabling the model to uniformly learn conversational gestures and rhythmic dance without explicit domain labels. Furthermore, experiments reveal a synergistic effect where the integration of diverse motion domains mutually reinforces generation fidelity.

Beyond real-time generation, our system is designed for practical deployment within modern voice agent ecosystems. It operates as a plug-and-play module, accepting audio streams from diverse sources ranging from web browsers to AI conversational platforms. Furthermore, we introduce a tool-call interface that enables upstream systems, such as Large Language Models, to interleave explicit semantic actions with implicit audio-driven motion, bridging the gap between purely reactive audio-driven animation and controllable, intent-driven behavior.

To further align real-time generation with human preferences, we explore Reinforcement Learning (RL) by investigating both reward-model-based strategies using Group Relative Policy Optimization (GRPO) and human-annotation-based strategies using Direct Preference Optimization (DPO). We demonstrate measurable improvements in perceived quality and provide an analysis of applying RL to online auto-regressive motion generation for future research.

Our primary contributions are: 
\begin{itemize}
    \item {A unified streaming architecture that leverages attention-based causal tokenization to synthesize continuous, high-fidelity motion from streaming speech and music with low latency.}
    \item A robust training curriculum utilizing Hierarchical Token Corruption to enable synergistic learning across diverse domains, boosting performance on individual tasks, alongside an exploration of RL strategies (GRPO/DPO) to enhance perceived generation quality.
    \item A deployable plug-and-play system that integrates with voice agents, supporting both implicit audio-driven animation and explicit semantic control via tool calls.
\end{itemize}

\section{Related Work}

\subsection{Co-speech Gesture Generation}
The trajectory of co-speech gesture generation reflects a fundamental paradigm shift from explicit, rule-based heuristics to implicit, data-driven synthesis. Early frameworks~\cite{kopp2006bml,cassell2001beat,lee2006nonverbal,lhommet2015cerebella,cassell1994rulefullbody} relied on rigid production rules and manual linguistic mappings, which offered controllability but lacked kinematic naturalness. The advent of deep learning initially spurred deterministic regression approaches~\cite{Gesticulator,liu2022beat,yoon2020speech,zhou2022GestureMaster,gesturematching22}; however, by modeling the modal average of plausible motions, these methods frequently suffered from ``mean-pose convergence'', resulting in over-smoothed and under-articulated output. To address the inherent stochasticity and one-to-many ambiguity of the speech-to-gesture mapping, the field has pivoted toward probabilistic generative modeling. This landscape encompasses Normalizing Flows~\cite{alexanderson2020style,ye2022audio} for explicit density estimation, Variational Autoencoders (VAEs)~\cite{ghorbani2022zeroeggs,li2021audio2gestures,ShiCVM2024} for continuous latent structuring, and Vector Quantized (VQ) frameworks~\cite{yazdian2022gesture2vec,ao2022rhythmic,ha2g:liu2022learning,liu2022videogeneration,talkshow:yi2022generating,Lu2023CoSpeechGS} that learn discrete motion codebooks. Diffusion Probabilistic Models based approaches~\cite{simon2023lda,Ao2023GestureDiffuCLIP,yang2023diffusestylegesture,cheng2024siggesture,zhang2024lmm,mughal2025retrieving,yang2025gesturehydra} excelling at modeling complex distributions via iterative denoising and MLLM~\cite{chen2025language, chen2025motionllm, hou2025motionverse, liu2025mag} unifying 3D human motion with text and speech in a shared latent space. 
{Among these, ConvoFusion~\cite{mughal2024convofusion} extends diffusion-based generation by enabling gesture emphasis on specific words. Teller~\cite{zhen2025teller} proposes a real-time audio-driven portrait talking head system. ACRNN~\cite{zhou2018arcnn} is the first method capable of generating arbitrarily long motions in real time with stability.}
While many approaches~\cite{chen2024diffsheg,liu2025gesturelsm} claim real-time capability for audio-driven body motion generation, they merely achieve generation speeds faster than playback speed under the assumption of full audio context availability. True streaming scenarios—where audio is incrementally received and motion is progressively generated—remain largely unexplored.

\subsection{Multimodal Motion Synthesis}
Motion synthesis research has significantly expanded its scope by integrating diverse control signals beyond audio. These range from semantic text descriptions~\cite{zhang2022motiondiffuse,tevet2023human,lu2025scamo,fan2025go,bae2025less} and spatial trajectory constraints~\cite{xie2023omnicontrol,wan2023tlcontrol,zheng2025autokeyframe} to physical interaction states~\cite{liu2025uni,lu2025choice,he2025syncdiff,ruiz2025mixermdm} and visual signals~\cite{feng2025physhmr,bekor2025gaussian}. Regarding stylistic control, motion examples~\cite{weiyu23GenMM,aberman2020unpaired} provide a direct reference for desired behaviors. While earlier methods like ZeroEGGS~\cite{ghorbani2022zeroeggs} compress these examples into static style vectors, often losing kinematic fidelity, recent approaches like MECo~\cite{chen2025meco} and PersonaBooth~\cite{kim2025personabooth} demonstrate that leveraging discrete token prefixes or personalized identifiers allows for precise, fine-grained control. However, within the specific domain of audio-driven generation, a critical fragmentation persists. While recent works achieve high fidelity in niche tasks, such as instrument-specific performance~\cite{qiu2025elgar}, complex rhythmic alignment~\cite{ghosh2025duetgen,nguyen2025learning}, {or training from in-the-wild short-form music-dance videos~\cite{zhao2024dancefusion}}, current systems are typically constrained to exclusive domains, handling either conversational speech or rhythmic dance. This bifurcation necessitates explicit task labels or separate models, {highlighting the lack of a framework capable of processing a unified, speech and music stream.}

\subsection{Reinforcement Learning}
Reinforcement Learning (RL) has long served as the standard paradigm for optimizing sequential decision-making~\cite{DBLP:books/lib/SuttonB98}, with policy gradient methods~\cite{williams1992simple, sutton1999policy, haarnoja2018soft} dominating high-dimensional motion control. However, applying these techniques to motion synthesis has historically been fraught with challenges. Prior approaches relying on offline RL~\cite{sunco, kumar2020conservative} or Actor-Critic frameworks~\cite{DBLP:conf/cvpr/SiyaoYGLW0L022} frequently struggled with brittle reward engineering~\cite{pinto2023tuning} and insufficient exploration. {Hybrid frameworks such as MotionVAE~\cite{ling2020motionvae} and AMDM~\cite{shi2024amdm} couple generative motion priors with RL-trained policy controllers to satisfy task-specific objectives.} In the generative era, the focus has shifted toward Reinforcement Learning from Human Feedback (RLHF)~\cite{ouyang2022training, menick2022teaching, yuan2023rrhf} to better capture perceptual nuance. To circumvent the well-documented instability of PPO-based pipelines, Direct Preference Optimization (DPO)~\cite{rafailov2023direct} has emerged as a robust alternative, optimizing policies directly from preference pairs without an explicit reward model—a strategy now proven across textual~\cite{she2024mapo, liu2024enhancing} and multimodal domains~\cite{zhou2024aligning, zhao2023beyond, li2023silkie}. Complementing this, Group Relative Policy Optimization (GRPO)~\cite{shao2024deepseekmath} introduces a mechanism for stable online refinement via group advantage normalization. We systematically explore these alignment strategies to improve perceived motion quality beyond the training distribution, particularly for robust one-shot streaming scenarios.

\section{Method}
\begin{figure}[t]
  \centering
  \includegraphics[width = \linewidth]{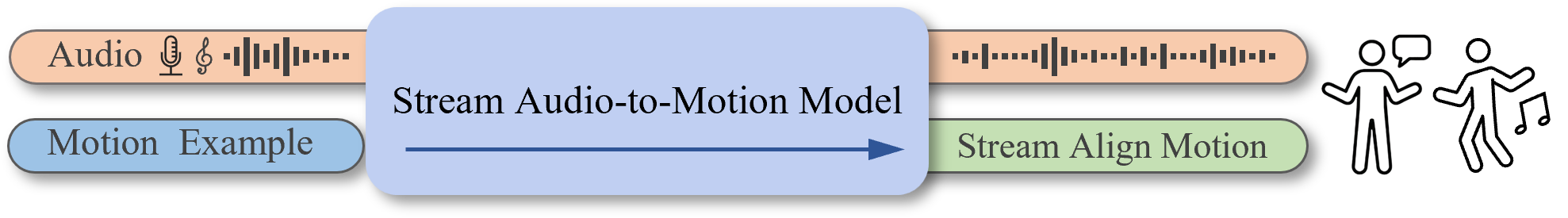}
  \caption{The structure of our motion generation model. Our model is capable of receiving streaming audio inputs and producing streaming motion outputs. Then, the time-aligned audio and motion are returned to the user together. Furthermore, our model can receive motion examples as additional  control signals.}
  \label{fig:overview}
\end{figure}

As depicted in Figure~\ref{fig:overview}, our framework is designed for real-time, high-fidelity 3D motion synthesis from streaming audio with low latency. Our system comprises three core components: first, a Causal Attention-based Motion Tokenizer that discretizes continuous motion manifolds into latent tokens without violating temporal causality; second, a repurposed pre-trained LLM generator optimized via a three-stage curriculum to align audio-motion modalities and enable explicit semantic control; and finally, a Reinforcement Learning (RL) Alignment stage that refines the policy to improve the alignment of generated motion with human perceptual standards for zero-retry streaming scenarios.

\subsection{Motion Tokenizer}
\label{subsec:motion_Tokenizer}

We define motion $\mathbf{m}_{1:N}$ as a sequence of pose states parameterized by root velocity, height, and 6D joint rotations~\cite{Zhou_2019_CVPR}. Standard discrete motion tokenizers typically rely on non-causal architectures that necessitate future-frame look-ahead, introducing latency that is prohibitive for real-time interaction. Although recent attempts~\cite{jiang2025causal,Xiao2025ICCV} enforce causality via convolutional left-padding. However, due to the limited expressive capacity of convolutional networks~\cite{Zhang2024SemanticGesture}, the reconstruction process often suffers from visual artifacts. To resolve this, we propose an Attention-based Causal Motion Tokenizer. We replace rigid convolutional backbones with stacked attention blocks governed by a causal mask, strictly confining the receptive field to the preceding $p$ frames. To handle temporal resampling without information loss, we adopt a dual-path strategy inspired by DC-AE~\cite{chen2024deep}. Downsampling is achieved by aggregating a temporal pooling branch with a feature concatenation branch processed via an MLP, while upsampling reconstructs temporal resolution through temporal replication combined with channel-expansion MLPs. Furthermore, to suppress physical artifacts such as foot sliding, we explicitly integrate Forward Kinematics (FK) into the optimization loop, imposing auxiliary losses on global joint positions, velocities, accelerations, and foot contact consistency. Detailed formulations are provided in the appendix.

The motion sequence $\mathbf{m}_{1:N}$ is encoded into a continuous latent trajectory $\mathbf{z}_{1:n} = \mathcal{E}(\mathbf{m}_{1:N})$, with a temporal downsampling ratio of $N/n$. To discretize this manifold, {we employ Residual Vector Quantization (RVQ)~\cite{zeghidour2021soundstream,guo2023momask,MoConVQ}}. The latent vector $\mathbf{z}$ is approximated as a summation of $Q$ quantized residuals, $\hat{\mathbf{z}} = \sum_{q=0}^{Q-1} \hat{\mathbf{z}}^q$, where each component $\hat{\mathbf{z}}^q$ is retrieved from a distinct codebook $\mathbf{C}_q$. This process is recursive: the initial layer quantizes the raw latent, while subsequent layers $q > 0$ refine the quantization error of the partial sum. The final discrete representation $\hat{\mathbf{z}}$ is decoded by $\mathcal{D}$ to reconstruct the motion $\hat{\mathbf{m}}$. The entire framework is optimized via a composite objective balancing kinematic reconstruction fidelity $\mathcal{L}_{\text{rec}}$ and codebook commitment:

\begin{equation}
\begin{split}
    \mathcal{L}_{\text{rec}} &= \|\hat{\mathbf{m}}_{1:N} - \mathbf{m}_{1:N}\|_1 + \eta \sum_{q=0}^{Q-1} \|\mathbf{z}^q_{1:n} - \operatorname{sg}[\hat{\mathbf{z}}^q_{1:n}]\|_2^2 \\
    &\quad + \Phi \Big( \operatorname{FK}(\hat{\mathbf{m}}_{1:N}), \operatorname{FK}(\mathbf{m}_{1:N}) \Big),
\end{split}
\end{equation}

where $\Phi$ encapsulates the FK-based auxiliary losses, $\operatorname{sg}[\cdot]$ denotes the stop-gradient operator, and $\eta$ weights the embedding constraint. To decouple part-specific dynamics, we implement anatomically partitioned tokenization, maintaining separate codebooks for the upper body, lower body, and hands.

\begin{table}[htbp]
    \centering
    \caption{Quantitative comparison results. MPJPE denotes the Mean Per-Joint Position Error computed in the character-centric coordinate system, measured in units of $10^{-4}\,\mathrm{m}$. Similarly, Trans Loss quantifies the average per-frame root  translation velocity error, also reported in $10^{-4}\,\mathrm{m}$.}
    \label{tab:methods_comparison}
    \resizebox{\linewidth}{!}{%
        \begin{tabular}{lcccc}
            \toprule
            \multirow{2}{*}{Methods} & \multicolumn{3}{c}{Reconstruction} & \multicolumn{1}{c}{Generation} \\
            \cmidrule(lr){2-4} \cmidrule(lr){5-5}
             & FID $\downarrow$ & MPJPE $\downarrow$ & Trans Loss $\downarrow$ & FID $\downarrow$ \\
            \midrule
            Real motion & 0 & 0 & 0 & 0 \\
            \midrule
            CausalConv-RVQ      & 9.208 & 525.6 & 12.94 & 18.06\\
            Attn                & 4.183 & 411.6 & 12.53 & 12.25 \\
            Attn(w/o dual)      & 12.21 & 982.4 & 48.77 & 18.55 \\
            Attn(w/o auxiliary) & 8.775 & 778.5 & 55.67 & 17.68 \\
            Attn(w/o lookback)  & 6.612 & 468.8 & 12.89 & 15.53\\
            Attn(w/ bodypart)   & 1.306 & 184.1 & 9.637 & 9.465  \\
            \bottomrule
        \end{tabular}%
    }
\end{table}

\begin{figure}[t]
  \centering
  \includegraphics[width=1.0\linewidth]{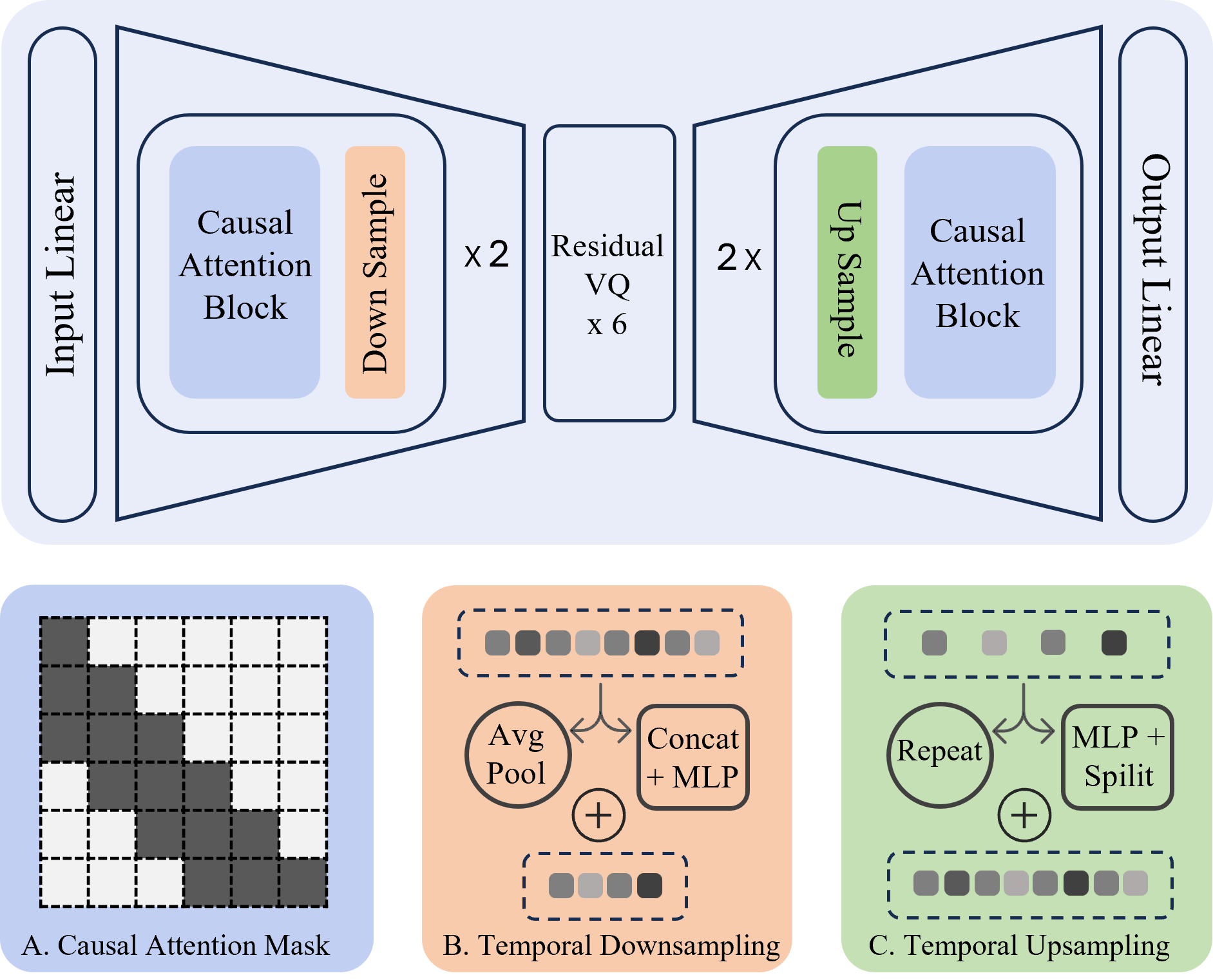}
\caption{Architecture of our Attention-based Causal Motion Tokenizer with Residual Vector Quantization. (A) Causal attention mask confines receptive field to preceding frames. (B) Temporal downsampling via dual-path aggregation. (C) Temporal upsampling via dual-path expansion.}
  \label{fig:2model}
\end{figure}

\subsection{Audio Driven Motion Generation}

Following the progressive learning paradigm in MECo~\cite{chen2025meco}, we orchestrate the adaptation of the pre-trained LLM for generative motion synthesis through a three-stage curriculum. This regimen systematically bridges the modality gap: (1) Embedding Space Alignment, which projects the discrete audio and motion codebooks into the LLM's continuous latent manifold; (2) Acoustic-Kinematic Alignment, which conditions the backbone to synthesize motion from streaming audio; and (3) Exemplar-Driven Control, which fine-tunes the model to accept reference motions as explicit stylistic directives.

To unify the input modalities, we employ the causal motion tokenizer detailed in Sec.~\ref{subsec:motion_Tokenizer} for kinematic discretization. For the acoustic modality—spanning both conversational speech and complex music—we utilize a causal variant of EnCodec~\cite{defossez2022encodec} to provide a consistent discrete interface.

Crucially, we diverge from MECo's strategy of prioritizing the primary quantization layer. We posit that high-fidelity reconstruction requires the explicit modeling of the full residual hierarchy. Consequently, we adopt the flattened interleaving strategy from MusicGen~\cite{copet2023musicgen}, serializing the multi-layer RVQ indices into a single autoregressive stream. To account for the non-uniform information density across quantization levels—where the initial layer captures fundamental dynamics and subsequent layers encode high-frequency residuals—we implement a Hierarchical Loss Scaling strategy. We apply monotonically decaying weights to the cross-entropy objectives of deeper RVQ layers, guiding the optimization to prioritize structural coherence before refining fine-grained details.





\subsubsection{Hierarchical Token Corruption}

We observe that unifying multiple audio-to-motion tasks within a shared motion token space induces a catastrophic failure mode: conditional collapse. As elucidated in our theoretical analysis (see Appendix), this pathology stems from a fundamental conflict in the optimization dynamics. The autoregressive motion prior remains strong in the learning signal, effectively ``short-circuiting'' the weaker audio conditioning, particularly when task-specific data is sparse. The model learns to aggregate next-motion-token probability more heavily on recent motion history while ignoring the acoustic input.

To counteract this, we propose Hierarchical Token Corruption, a targeted regularization strategy designed to recalibrate these dynamics. By stochastically perturbing context motion tokens during training, we actively penalize over-reliance on the autoregressive history and force the model to rely on the mutual information between the audio condition and the target motion.

Unlike uniform noise injection, our perturbation strategy respects the structural hierarchy of Residual Vector Quantization (RVQ). For each timestep selected for corruption, we sample a layer depth $\ell_t \sim \text{Uniform}(1, L)$ and randomize tokens from layer $\ell_t$ through $L$, while leaving the coarser, foundational layers intact. This approach yields two critical benefits. First, it mimics realistic generation artifacts—where fine-grained details degrade before global structure—thereby serving as a robust data augmentation technique. Second, it instills error-correcting capabilities; the model learns to recover ground-truth trajectories even when conditioned on perturbed context, ensuring graceful recovery from sampling errors during long-form autoregressive inference.

\subsubsection{Example Control}
Following MECo, we integrate exemplar-based control to steer generation. While our generative backbone models the full Residual Vector Quantization (RVQ) hierarchy to maximize fidelity, we observe that the semantic density of the motion signal is predominantly concentrated in the primary VQ layer. Consequently, we constrain the conditioning mechanism to extract control tokens exclusively from the first-level codebook of the reference sequence. 


\subsection{Reinforcement Learning}

To enhance generation quality beyond the training distribution, we employ Reinforcement Learning (RL) to align the model's policy with human perceptual standards. We investigate two paradigms: Reward-Guided Optimization (leveraging self-supervised proxy rewards) and Direct Preference Alignment (leveraging human feedback).

\subsubsection{Reward-Guided Optimization (GRPO)}
In scenarios lacking explicit human labels, we synthesize a proxy reward signal combining intrinsic motion fidelity and cross-modal synchronization. We optimize this objective using Group Relative Policy Optimization (GRPO)~\cite{shao2024deepseekmath}, which stabilizes training by normalizing advantages within a sampled group. The objective is formulated as:
\begin{equation}
\mathcal{L}_{\text{GRPO}} = -\frac{1}{G} \sum_{i=1}^{G} \rho_i \hat{A}_i + \beta_{G} \, \mathbb{D}_{\text{KL}}(\pi_\theta \| \pi_{\text{ref}}),
\end{equation}
where $\rho_i$ denotes the importance ratio $\pi_\theta(y_i|x) / \pi_{\theta_{\text{old}}}(y_i|x)$, and $\hat{A}_i = (r_i - \mu_G) / \sigma_G$ represents the group-normalized advantage. The Kullback-Leibler (KL) divergence term ensures the optimized policy $\pi_\theta$ does not deviate excessively from the reference policy $\pi_{\text{ref}}$.

\paragraph{Self-Supervised Motion Quality Reward.}
Constructing a robust quality metric without manual annotation is non-trivial. {Inspired by the degradation modeling in E3D2~\cite{wang2024e3d2}, UnifiedGesture~\cite{yang2023unifiedgesture} and D-REX~\cite{brown2019drex}, we establish a self-supervised quality curriculum by artificially corrupting ground-truth motion sequences. We apply variable rates of random and Hierarchical Token Corruption to generate a synthetic dataset with known degradation levels, calibrated via FID scores. This establishes a monotonic mapping between corruption severity and quality}, which serves as the training signal for our reward model. The reward model architecture mirrors our motion tokenizer but utilizes bidirectional attention to capture global temporal context and omits the quantization layer to output continuous quality scalars.

\paragraph{Audio-Motion Alignment Reward.}
To evaluate rhythmic alignment, we learn a joint multimodal embedding space using the InfoNCE contrastive objective~\cite{oord2018representation,radford2021learning}. We utilize the pre-trained BEATs~\cite{chen2023beats} model as the audio encoder and a randomly initialized Transformer as the motion encoder. The reward is defined as the cosine similarity between the synchronized audio and motion embeddings, encouraging the policy to maximize cross-modal coherence.

\subsubsection{Direct Preference Alignment (DPO)}
When human feedback is available, we bypass proxy reward modeling and optimize the policy directly against human preferences using Direct Preference Optimization (DPO)~\cite{rafailov2023direct}. This approach implicitly solves the reward maximization problem without the instability of a separate reward network:
\begin{equation}
\mathcal{L}_{\text{DPO}} = -\mathbb{E} \left[ \log \sigma \left( \beta_D \log \frac{\pi_\theta(y_w | x)}{\pi_{\text{ref}}(y_w | x)} - \beta_D \log \frac{\pi_\theta(y_l | x)}{\pi_{\text{ref}}(y_l | x)} \right) \right],
\end{equation}
where $y_w$ and $y_l$ denote the preferred (winning) and dispreferred (losing) motion sequences, respectively, and $\beta_D$ modulates the strength of the KL constraint. To construct the preference dataset $\mathcal{D}$, we employ a Best-of-N sampling strategy: for each audio input, we generate eight candidate sequences using the pre-trained model. Human annotators perform a comparative evaluation to identify the optimal and least plausible samples, forming $(y_w, y_l)$ pairs. To ensure high signal-to-noise ratio in the preference data, pairs lacking a distinct quality disparity are filtered out.

\section{Experiment}
\subsection{Datasets and Preprocessing}
To evaluate our framework across disparate kinematic domains, we leverage two complementary high-fidelity datasets: ZeroEGGS, a stylized speech-gesture corpus (approx. 2 hours) capturing a single speaker across 19 distinct expressive styles; and Motorica, a rhythmic dance database (approx. 6 hours) featuring five performers across eight diverse genres. To resolve topological discrepancies between sources, we adopt the standardized skeletal representation proposed by Holden~\cite{zeroeggs_retarget,motorica_retarget}. Subsequently, we employ kinematic retargeting (Autodesk Maya) to transfer all motion data onto a unified target digital character.

\paragraph{RL Alignment Corpus.} To facilitate reinforcement learning beyond the constraints of the original paired dataset, we curate a supplementary collection of unannotated audio. For the speech domain, we synthesize approximately one hour of conversational dialogue using Gemini 3 Pro scripts rendered via ElevenLabs' neural TTS. For the music domain, we assemble a diverse corpus of 100 compositions from open platforms (YouTube), covering a broad spectrum of tempos and genres to enhance rhythmic generalization.

\paragraph{Motion Refinement.} 
We observe that a significant portion of the Motorica dataset exhibits artifacts characterized by erratic or static finger motion. To rectify this, we train a motion inpainting model~\cite{shafir2024humanprior} leveraging the high-fidelity finger motion data from ZeroEGGS. This model synthesizes plausible finger motion conditioned on the remaining body joints. A Savitzky-Golay filter is then applied to the synthesized finger motion to mitigate temporal jitter.

\paragraph{Facial Animation.} To animate our digital avatar, we utilize the industry-standard Apple ARKit blendshape schema. We curated a proprietary facial capture dataset (approx. 1 hour) consisting of synchronized speech and high-fidelity blendshape weights. The acquisition pipeline utilized an iPhone 12 running Live Link Face (Epic Games, Inc.), with the actor performing the Harvard Sentences~\cite{IEEE1969} corpus to ensure comprehensive phonemic coverage. Based on our collected data, we train a lightweight streaming speech-to-facial animation model following~\cite{chen2025dystream}. More details are provided in the Appendix.

\begin{table}[htbp]
    \centering
    \caption{Quantitative evaluation on test set. We report $BA_G$ $\times 10^{-1}$, $BA_D$ $\times 10^{-1}$. \textbf{Bold} face indicates the best result. {"Ours" denotes the no-RL model.}}
    \label{tab:zm_comparison}
    \setlength{\tabcolsep}{4pt} 
    \begin{tabular}{l c c c c}
        \toprule
        Method & FID $\downarrow$ & Diversity $\uparrow$ & $\text{BA}_\text{G}$ $\uparrow$ & $\text{BA}_\text{D}$ $\uparrow$ \\
        \midrule
        GT    & 0 & 21.52 & 7.775 & 2.619 \\
        MECo     & 14.73     & 23.13 & 7.507 &  2.622 \\
        EDGE   & 18.06 & 19.71 & 8.190 & \textbf{2.668}  \\
        Ours (w/o corrupt)    & 25.92 & \textbf{29.58} & \textbf{8.464}  & 2.541\\
        Ours   & \textbf{9.465} & 20.70 & 8.277 & 2.603 \\
        \midrule
        Ours (DPO)   & 12.39 & 19.67 & 8.283 & 2.607 \\
        Ours (GRPO)   & 24.13 & 20.89 & 8.239 & 2.618 \\
        \bottomrule
    \end{tabular}
\end{table}

\begin{table*}[t]
\centering
\caption{User Study Results. We evaluate our method on both Dance and Gesture generation tasks across three comparative settings. The metrics reported are Human Likeness, Beat Matching, and Overall Preference. All results are presented as \textit{mean} $\pm$ \textit{95\% confidence interval}. {The three categories are independent.} }
\label{tab:user_study}

\renewcommand{\arraystretch}{1.2}

\resizebox{\textwidth}{!}{%
\begin{tabular}{ll ccc c ccc}
\toprule
\multirow{2.5}{*}{Category} & \multirow{2.5}{*}{Method} &
\multicolumn{3}{c}{Dance} & & \multicolumn{3}{c}{Gesture} \\
\cmidrule(lr){3-5} \cmidrule(lr){7-9}

 & & Human Likeness & Beat Matching & Overall Preference & &
     Human Likeness & Beat Matching & Overall Preference \\
\midrule

\multirow{3}{*}{Comparison with SOTA}
& MECo & $-0.508 \pm 0.244$ & $-0.277 \pm 0.230$ & $-0.477 \pm 0.236$ & & $0.235 \pm 0.198$ & $0.061 \pm 0.222$ & $0.096 \pm 0.216$ \\
 & EDGE & $0.148 \pm 0.238$ & $-0.136 \pm 0.174$ & $0.099 \pm 0.229$ & & $-0.676 \pm 0.154$ & $-0.705 \pm 0.149$ & $-0.748 \pm 0.143$ \\
 & Ours         & $0.244 \pm 0.237$ & $0.337 \pm 0.188$ & $0.267 \pm 0.234$ & & $0.588 \pm 0.150$ & $0.798 \pm 0.155$ & $0.816 \pm 0.137$ \\

\midrule
\multirow{4}{*}{RL Strategy Ablation}
 & {Ours}         & $-0.078 \pm 0.156$ & $-0.028 \pm 0.137$ & $-0.085 \pm 0.170$ & & $-0.231 \pm 0.262$ & $0.000 \pm 0.219$ & $-0.169 \pm 0.269$ \\
 & {Ours (DPO)}     & $0.078 \pm 0.156$ & $0.028 \pm 0.137$ & $0.085 \pm 0.170$ & & $0.231 \pm 0.262$ & $0.000 \pm 0.219$ & $0.169 \pm 0.269$ \\
\cmidrule{2-9}
 & {Ours}     & $-0.109 \pm 0.215$ & $-0.069 \pm 0.188$ & $-0.188 \pm 0.211$ & & $-0.109 \pm 0.193$ & $-0.092 \pm 0.162$ & $-0.059 \pm 0.191$ \\
& {Ours (GRPO)}    & $0.109 \pm 0.215$ & $0.069 \pm 0.188$ & $0.188 \pm 0.211$ & & $0.109 \pm 0.193$ & $0.092 \pm 0.162$ & $0.059 \pm 0.191$ \\
\midrule

\multirow{3}{*}{Dataset Composition}
 & Gesture Only       & - & - & - & & $-0.345 \pm 0.192$ & $-0.727 \pm 0.193$ & $-0.555 \pm 0.198$ \\
 & Dance Only         & $-0.350 \pm 0.141$ & $-0.355 \pm 0.124$ & $-0.323 \pm 0.136$ & & - & - & - \\
 & Merged & $0.350 \pm 0.141$ & $0.355 \pm 0.124$ & $0.323 \pm 0.136$ & & $0.345 \pm 0.192$ & $0.727 \pm 0.193$ & $0.555 \pm 0.198$ \\

\bottomrule
\end{tabular}%
}
\end{table*}

\begin{table}[t]
\centering

\caption{Comparison with the state-of-the art methods on BEAT2~\cite{liu2024emage} test set. Quantitative evaluation on BEAT2. We report {FID} $\times 10^{-1}$, {$\text{BA}_\text{G}$} $\times 10^{-1}$, and diversity. \textbf{Bold} face indicates the best result.}
\label{tab:beat2_benchmark}
{
\begin{tabular}{lccc}
\toprule
     Method & {FID} $\downarrow$ & {$\text{BA}_\text{G}$} $\uparrow$ & Diversity~$\uparrow$   \\ 

\midrule
S2G\cite{ginosar2019learning} & 28.15  & 4.683  & 5.971                        \\
Trimodal\cite{yoon2020speech} & 12.41  & 5.933  & 7.724                         \\
HA2G\cite{ha2g:liu2022learning} & 12.32  & 6.779 & 8.626                        \\
DisCo\cite{liu2022disco} & 9.417  & 6.439 &  9.912                              \\
CaMN\cite{liu2022beat} & 6.644  & 6.769 & 10.86                                  \\
DiffStyleGesture\cite{yang2023diffusestylegesture} & 8.811 & 7.241 & 11.49     \\
Habibie \textit{et al}.\cite{habibie2021learning} & 9.040 &  7.716 &  8.213       \\
TalkShow\cite{talkshow:yi2022generating} &  6.209 &  6.947 & 13.47                \\      
EMAGE \cite{liu2024emage} & 5.512 &  7.724 & 13.06                               \\
SynTalker\cite{chen2024syntalker}&  6.413  & {7.971} & 12.72                      \\ 
MECo \cite{chen2025meco} & 3.401 & 7.346  & \textbf{15.30} \\
ViBES \cite{zhang2026vibes}& 5.257 &\textbf{8.103}& 13.03 \\
PersonaGesture \cite{zhang2026personagesture}& 3.930& 7.100 &13.25\\
Ours & \textbf{2.874} & 7.342  & 13.53 \\
\bottomrule
\end{tabular}}
\label{table1}
\end{table}

\subsection{Settings}

Our system synthesizes native motion at 30 frames per second (FPS), which is subsequently interpolated to 60 FPS for final rendering. We detail the configuration for each component. The RVQ-VAE (Sec.~\ref{subsec:motion_Tokenizer}) is trained with a temporal downsampling factor $n/N = 4$, yielding a latent motion rate of 7.5 Hz. We employ a codebook size $K = 512$, latent dimension $d = 512$, and quantization depth $Q = 6$. The model is optimized using a batch size of 256, a commitment loss weight $\eta = 0.1$, and a learning rate of $4 \times 10^{-4}$ managed by a step decay scheduler. During training, we randomly sample 64-frame motion windows. We adopt Qwen2.5-0.5B-Instruct~\cite{qwen2.5} as the base generator, detaching its tied input/output embeddings to accommodate our modality-specific vocabularies. The model processes 4-second context windows, comprising 600 audio tokens {(derived from the first 2 RVQ layers of EnCodec at 75Hz)} and 540 motion tokens (flattened across 6 RVQ layers for three body partitions). Fine-tuning is performed with a batch size of 256 and a learning rate of $5 \times 10^{-5}$. For the reinforcement learning stage, we reduce the batch size to 16 and adjust the learning rate to $6 \times 10^{-5}$. In the GRPO configuration, we set the KL penalty $\beta_G = 0.01$ and perform 30 rollouts per prompt. For DPO, we utilize a deviation penalty $\beta_D = 0.1$. All experiments are conducted on a node equipped with two NVIDIA H200 GPUs. The complete training pipeline requires approximately 30 hours. At inference, our optimized pipeline achieves a throughput of $\sim$300 tokens/s, well within the latency budget for real-time interaction.

\subsection{Real-time Deployment}
\begin{figure*}
  \centering
  \includegraphics[width = \linewidth]{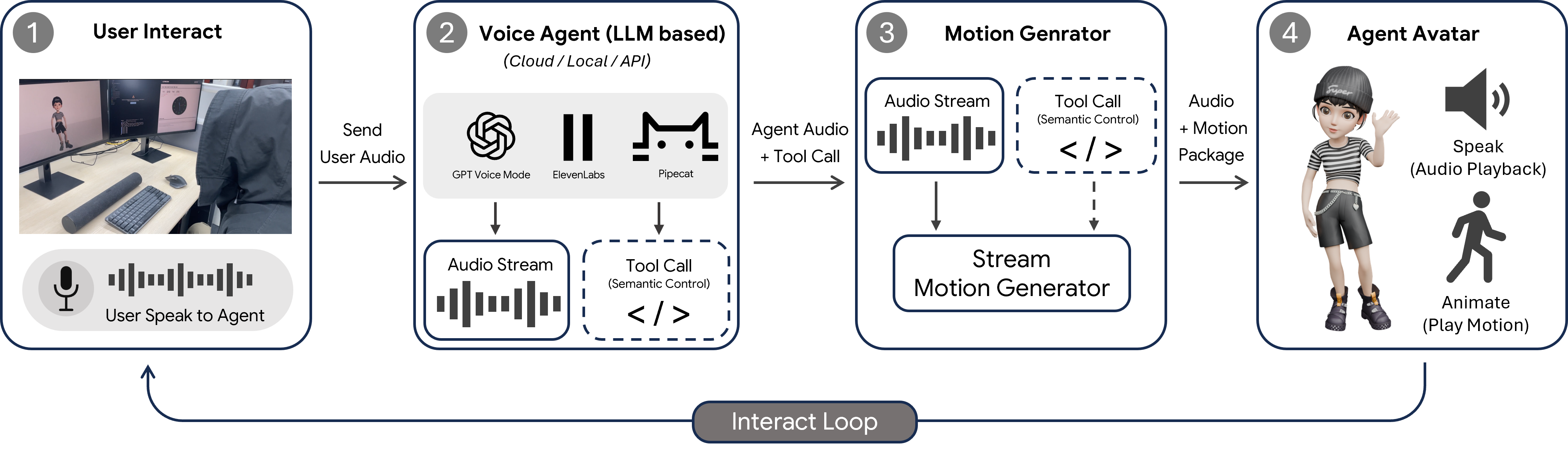}
  \caption{{Real-time Deployment. Our system comprises three components: the user host machine, a voice agent, and the Motion Generator. The host machine captures the user's voice via microphone (1) and streams it to the voice agent (2). The voice agent can be an omni-model (e.g., OpenAI's GPT voice mode) or a cascaded pipeline of VAD, ASR, LLM, and TTS modules (e.g., ElevenLabs, Pipecat), and can be deployed in the cloud, run locally, or accessed via API. It outputs an audio stream and, when appropriate, emits semantic control signals through a tool-call interface. The Motion Generator (3) consumes the audio stream and synchronously produces a motion stream, optionally conditioned on a motion example retrieved via the semantic control signal. The time-aligned audio and motion are then packaged and sent to the Rendering Client Frontend on the host machine to drive and visualize the avatar (4), closing the interaction loop.}}

  \label{fig:demo_pipeline}
\end{figure*}

{To enable user-friendly interactive avatars, as shown in ~\autoref{fig:demo_pipeline}, we build a distributed system composed of three functional tiers: a cloud-hosted Conversational Voice Agent (ElevenLabs), a rendering Client Frontend, and a dedicated GPU Inference Server.} We achieve continuous autoregressive streaming by deploying our fixed-context trained model via a sliding window strategy, generating motion in granular steps of 0.266 seconds (8 frames). We leverage CUDA Graph instantiation to reduce kernel scheduling overhead. Latency profiling across four processing stages (see appendix), including Audio Encoding, Motion Synthesis, Motion Decoding, and IK Post-processing, confirms that our total computational latency remains well below the 266ms audio chunk duration on both NVIDIA H200 and RTX 4090 platforms.


\subsection{Subjective Evaluation Protocol}
Following established subjective evaluation standards~\cite{Ao2023GestureDiffuCLIP,simon2023lda}, we assess generation quality across three perceptual dimensions: Human Likeness, Rhythmic Synchronization (Beat Matching), and Overall Preference. We adopt a rigorous pairwise comparison protocol: for each trial, participants are presented with two sequential 10-second clips synthesized by competing models conditioned on identical audio inputs. Evaluators indicate both the direction and intensity of their preference on a 5-point Likert scale (0: Neutral, 2: Strong Preference). To facilitate quantitative analysis, these ordinal ratings are mapped to a symmetric interval $[-2, 2]$, where positive values signify a preference for our method. The final subjective score is aggregated from 1,680 individual pairwise judgments, ensuring statistical significance.

\subsection{Quantitative Benchmarking} 
Given our framework's unified capability, we evaluate performance across both speech-to-gesture and music-to-dance domains. We standardize the measurement of distribution fidelity (FID) and generative Diversity across tasks. For rhythmic alignment, we employ domain-specific heuristics to capture the distinct temporal dynamics of each modality: for speech-gesture alignment, following EMAGE, we quantify the synchronization between acoustic onsets and the local minima of kinematic velocity; for music-dance alignment, following~\cite{davis2018beat}, we assess the correspondence between musical beats and the local maxima of motion deceleration (Detailed formulations for all metrics are provided in the Appendix).

We benchmark against leading domain-specific baselines: MECo for co-speech gesture and EDGE for music-driven dance. As summarized in Table~\ref{tab:zm_comparison} and Table~\ref{tab:user_study}, our unified approach consistently surpasses these specialized baselines in both objective metrics and subjective preference. Furthermore, evaluations on the high-fidelity BEAT2 benchmark (Table~\ref{tab:beat2_benchmark}) confirm that our method establishes a new state-of-the-art in generative fidelity (FID).

\subsection{Ablation Study}

\subsubsection{Attention-based Causal Motion
Tokenizer} We validate our motion tokenizer through four ablation studies and two controlled comparisons, assessing both intrinsic reconstruction fidelity and downstream generation efficacy. To quantify reconstruction quality, we report FID, MPJPE, and a Translation Loss (\emph{Trans Loss}), defined as the deviation between predicted and ground-truth root velocities. To assess downstream impact, we evaluate the FID of an audio-to-motion generator trained atop each tokenizer variant. Across all experiments, the inclusion of each proposed component yields consistent improvements in both signal reconstruction and generative quality. Furthermore, to ensure rigorous benchmarking, we compare against a CausalConv baseline implemented with an identical RVQ configuration and loss landscape; our attention-based approach demonstrates superior performance across all metrics.


\subsubsection{Hierarchical Token Corruption} 
\begin{figure*}[t]
  \centering
  \includegraphics[width = \linewidth]{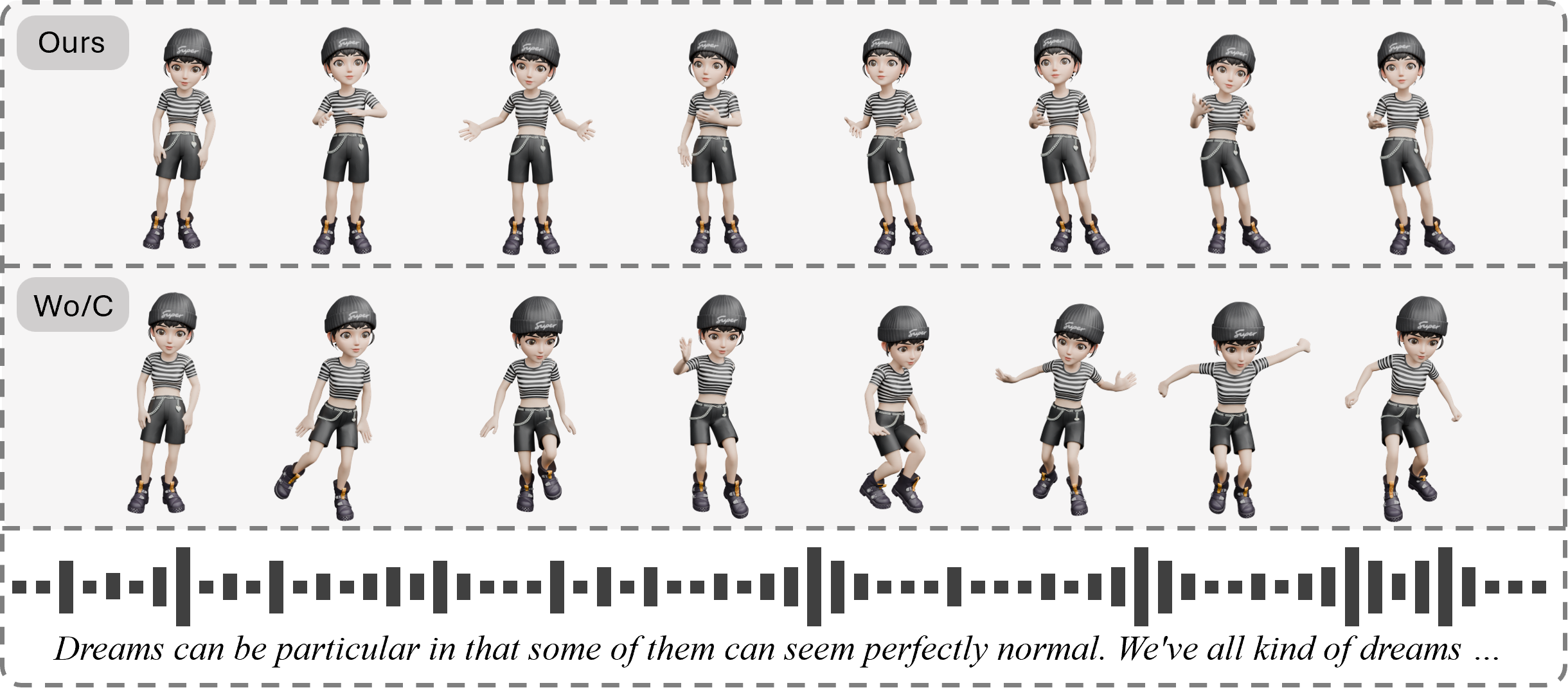}
  \caption{W/o C denotes training without Hierarchical Token Corruption. Given the same audio and initial motion input, our method generates natural motions that are well-synchronized with the audio. In contrast, the variant trained without Hierarchical Token Corruption largely ignores the audio input and produces erratic, dance-like motions that lack proper audio-motion correspondence.}
  \label{fig:fig_cmp_wocorrupt_2}
\end{figure*}

We identify Hierarchical Token Corruption as the linchpin of our unified training strategy. As illustrated in~\autoref{tab:zm_comparison} and~\autoref{fig:fig_cmp_wocorrupt_2}, ablating this mechanism leads to severe conditional collapse: the model ignores the input condition and persistently generates meaningless, physically implausible dance-like motions even during silence or neutral speech. Paradoxically, this pathological behavior results in the highest scores for Diversity and $BA_G$, as the ungrounded, high-variance movements artificially inflate these metrics without reflecting genuine perceptual quality. By reintroducing our hierarchical corruption strategy, the model successfully learns to adhere to the acoustic signal, enabling label-free learning from the joint dataset. Moreover, the corruption-augmented model achieves superior performance on individual tasks compared to single-task baselines, demonstrating that it effectively learns from cross-task training data.

\subsubsection{Cross-Modal Synergy via Joint Training} 
\begin{figure*}[t]
  \centering
  \includegraphics[width = \linewidth]{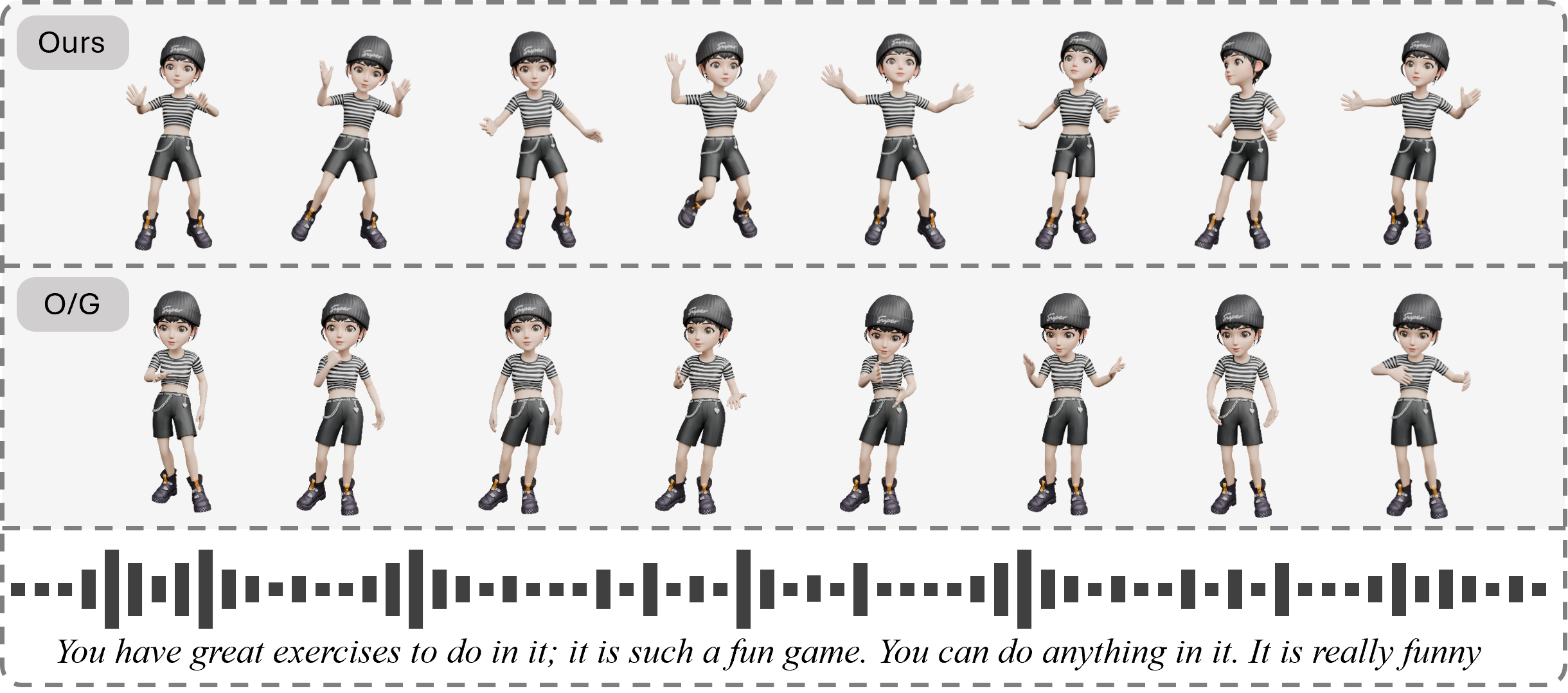}
  \caption{O/G denotes Gesture Only, which training exclusively on the speech-gesture dataset. As shown, our model trained jointly on both speech-gesture and music-dance datasets can produce exuberant, dance-like movements in response to cheerful audio, demonstrating its ability to generalize across motion domains and adapt motion style to audio characteristics.}
  \label{fig:fig_cmp_onlyzeroeggs_1}
\end{figure*}

We further investigate the efficacy of dataset composition by comparing three training configurations: Gesture-Only, Dance-Only, and Combined. As detailed in Table~\ref{tab:user_study}, joint training yields a performance uplift across both domains. Most notably, the inclusion of the music-to-dance dataset significantly enhances the beat-matching capability of the gesture generation. We attribute this to cross-modal synergy: the model internalizes robust rhythmic priors from the highly structured dance data and transfers this sensitivity to the speech domain. This transfer is particularly vital for gesture subsets with sparse rhythmic cues (e.g., ``Still'' or ``Flirty'' styles), where the speaker exhibits low kinematic variance. Furthermore, we observe an emergent zero-shot stylistic transfer: as shown in ~\autoref{fig:fig_cmp_onlyzeroeggs_1}, when driven by highly energetic ``happy'' speech, the agent occasionally produces lively, rhythmic gestures that were not present in the original speech dataset. This suggests that our unified framework possesses a degree of semantic generalization, mapping audio features to motion primitives regardless of the source domain.

\subsubsection{Reinforcement Learning Strategy Analysis} We conduct a comparative analysis of two alignment strategies: Direct Preference Optimization (DPO), utilizing human preference labels, and Group Relative Policy Optimization (GRPO), utilizing proxy rewards. Table~\ref{tab:user_study} confirms that both methods successfully align the model with human perceptions, improving subjective ratings over baseline.

\begin{figure*}
  \centering
  \includegraphics[width = 0.9\linewidth]{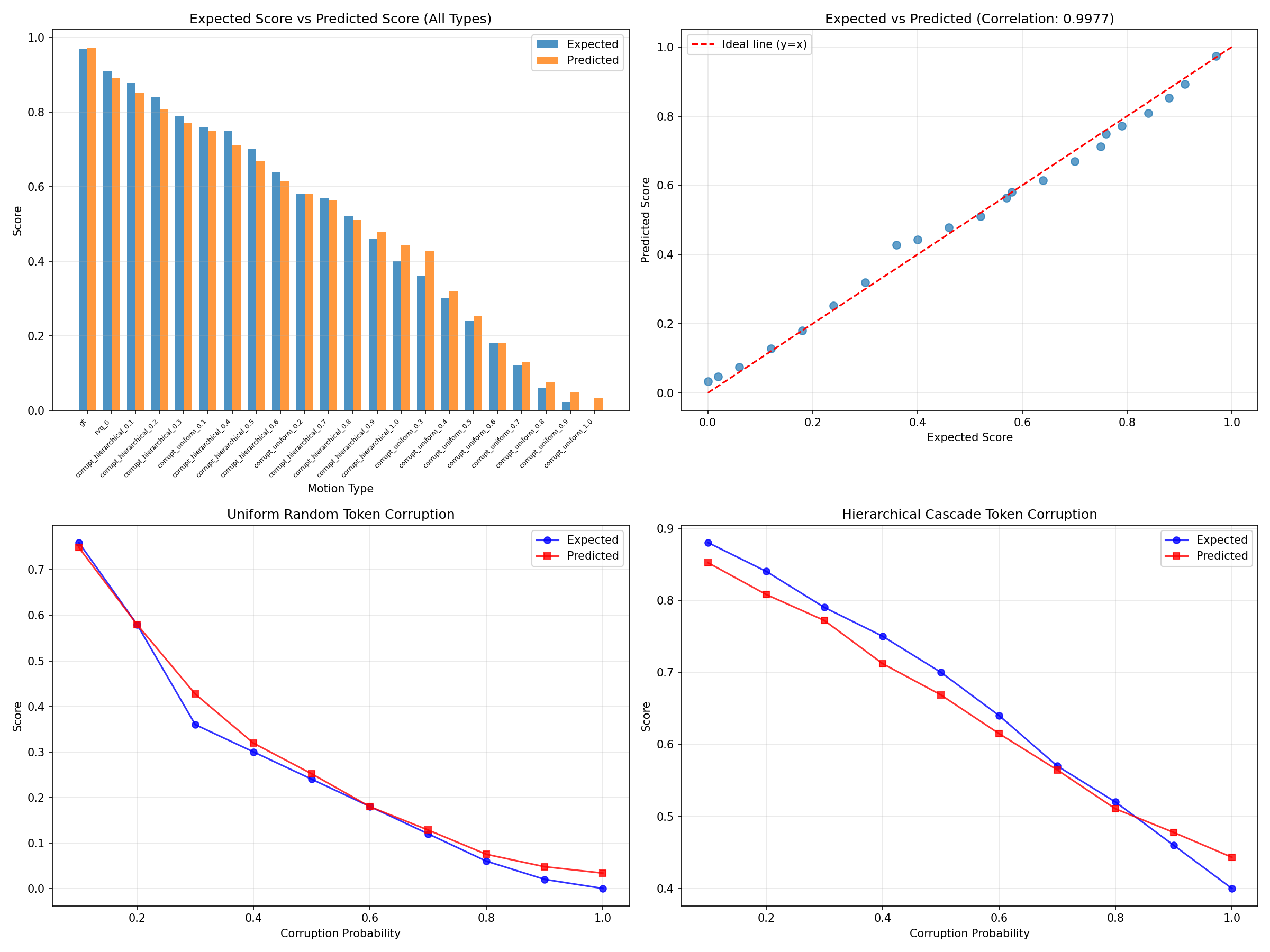}
  \caption{Motion Quality Reward Model Evaluation. The four plots demonstrate the performance of our motion quality reward model on the validation set under different corruption strategies. The visualization shows that our reward model exhibits strong generalization across various types of motion degradation.}
  \label{fig:motion_reward_model}
\end{figure*}

\paragraph{Reward Model Efficacy.} To validate the proxy signals used in GRPO, we evaluate our trained reward models on held-out test data. As depicted in Fig.~\ref{fig:motion_reward_model}, the motion quality model exhibits a Pearson correlation of $0.9977$ with ground-truth degradation levels, correctly preserving the ordinal ranking of motion quality across all corruption intensities. We also tested audio-motion alignment reward with retrieval metrics following the evaluation protocol of TMR~\cite{petrovich23tmr}, which achieves a retrieval success rate approximately $100\times$ higher than random chance, confirming its discriminative effectiveness. Please see appendix for details.

\paragraph{The Alignment-Fidelity Trade-off.} Despite the robustness of our reward models, Table~\ref{tab:zm_comparison} reveals a characteristic trade-off: both RL strategies induce a degradation in FID scores, with GRPO exhibiting a more pronounced divergence ($9.465 \to 24.13$) compared to DPO ($9.465 \to 12.39$). ~{This outcome is consistent with prior observations that reward optimization induces mode-seeking behavior: the policy concentrates mass on high-reward modes, which reduces distributional coverage (and thus inflates FID) while improving alignment with the target reward and human preference~\cite{ouyang2022instructgpt}.}

\paragraph{Domain-Specific Strategy Selection.}
Our user study reveals divergent effectiveness across motion domains. GRPO achieves stronger preference improvements on dance (Overall: $+0.188$ vs. DPO's $+0.085$), while DPO outperforms GRPO on gesture ($+0.169$ vs. $+0.059$). We attribute this to domain characteristics: dance motion favors strong rhythmic synchronization and tolerates exaggerated movements (or even benefits from them), aligning well with GRPO's aggressive optimization. Conversely, conversational gestures prioritize subtlety and naturalness, which are better preserved by DPO's conservative, preference-based learning.

\section{Discussions and Future Work} 
While this work establishes a robust baseline for unified real-time animation, several frontiers remain for future investigation. First, our current architecture decouples facial and body dynamics and lacks detailed non-verbal interaction modeling such as gaze, limiting the holistic cohesion required for deep engagement. {Second, reliance on acoustic features combined with limited speaker diversity in our training data can lead to domain confusion, such as misidentifying male speech as musical vocals and erroneously generating dance motions. Furthermore, the system currently lacks specific transition policies for abrupt acoustic terminations, leading to unnatural motion especially when music stops suddenly.} Finally, our framework focuses exclusively on the speaker role, neglecting the reciprocal nature of dyadic communication. Realizing true embodied interaction necessitates extending our generative paradigm to support active listening, enabling the avatar to synthesize non-verbal backchannels and reactive behaviors in response to user input~\cite{Ng_2022_CVPR} or environmental context~\cite{xu2025mospa}.

{
\begin{acks}
We are grateful to Linzhou Li for refining the teaser image, and to Jiacheng Guo and Yixuan Lai for their extensive efforts in manually evaluating and annotating the generated results. This work is partially supported by NSF China (No. 62572430, 62421003) and the XPLORER PRIZE.
\end{acks}
}

\bibliographystyle{ACM-Reference-Format}
\bibliography{sample-base}

\clearpage

\clearpage
\appendix

\section{More Details on Real-time Deployment}

\begin{table*}[t]
\centering
\caption{Experiments on retrieval ability of audio-motion contrastive space. Details are in Sec.~\ref{sec:reward_computation}}
\label{tab:audio_reward_model}
\renewcommand{\arraystretch}{1.2}
\setlength{\tabcolsep}{4pt}
\resizebox{\textwidth}{!}{
\begin{tabular}{ll|cccccc|cccccc}
\toprule
\multirow{2}{*}{\textbf{Protocol}} & \multirow{2}{*}{\textbf{BaseModel}} & \multicolumn{6}{c|}{Audio-motion retrieval} & \multicolumn{6}{c}{Motion-audio retrieval} \\
 & & R@1$\uparrow$ & R@3$\uparrow$ & R@5$\uparrow$ & R@10$\uparrow$ & MedR$\downarrow$ & MRR$\uparrow$ & R@1$\uparrow$ & R@3$\uparrow$ & R@5$\uparrow$ & R@10$\uparrow$ & MedR$\downarrow$ & MRR$\uparrow$ \\
\midrule
(a) All & \textbf{BEATs} & \textbf{6.42} & \textbf{10.51} & \textbf{14.60} & \textbf{21.79} & \textbf{67.0} & \textbf{11.07} & \textbf{6.64} & \textbf{10.51} & \textbf{13.77} & \textbf{18.75} & \textbf{75.0} & \textbf{10.70} \\
($N=1808$) & \textbf{Wav2CLIP} & 3.60 & 6.58 & 9.13 & 13.05 & 169.0 & 7.00 & 3.48 & 5.86 & 8.52 & 12.06 & 184.0 & 6.57 \\
 & \textbf{Random} & 0.06 & 0.17 & 0.28 & 0.55 & 904.0 & 0.41 & 0.06 & 0.17 & 0.28 & 0.55 & 904.0 & 0.41 \\
\midrule
(b) Small batches & \textbf{BEATs} & \textbf{18.67} & \textbf{27.33} & \textbf{33.33} & \textbf{44.00} & \textbf{15.0} & \textbf{26.49} & \textbf{21.67} & \textbf{30.00} & \textbf{34.67} & \textbf{44.00} & \textbf{13.0} & \textbf{28.37} \\
($N=300$) & \textbf{Wav2CLIP} & 9.00 & 13.67 & 17.67 & 28.67 & 23.5 & 15.69 & 6.33 & 11.67 & 16.00 & 24.67 & 37.5 & 12.14 \\
 & \textbf{Random} & 0.33 & 1.00 & 1.67 & 3.33 & 150.0 & 2.09 & 0.33 & 1.00 & 1.67 & 3.33 & 150.0 & 2.09 \\
\bottomrule
\end{tabular}
}
\end{table*}

\begin{table}[htbp]
\centering
\caption{Real-time Performance Evaluation. We report the mean and standard deviation of latency (ms) for each processing stage, averaged over 20 intermediate inference steps. Audio is processed in 266ms chunks.}
\label{tab:realtime_performance}
\begin{tabular}{lcc}
\toprule
{Processing Stage} & {NVIDIA H200} & {NVIDIA RTX 4090} \\
\midrule
Audio Encoder & $51.155 \pm 0.692$ & $64.041 \pm 5.062$ \\ 
Audio-to-Motion Model & $102.473 \pm 0.700$ & $118.932 \pm 3.428$ \\
Motion Decoder & $1.532 \pm 0.057 $ & $1.386 \pm 0.064$ \\
IK Post-processing & $13.990 \pm 0.940$ & $20.86 \pm 2.520$ \\
\midrule
Total Latency & $177.426 \pm 1.567$ & $215.823 \pm 4.887$ \\
\bottomrule
\end{tabular}
\end{table}
\subsection{Distributed System Topology} To enable high-fidelity interactive avatars, we architect a distributed system composed of three functional tiers: the Conversational Agent, the Client Frontend, and the Inference Server. The conversational Agent, hosted on the ElevenLabs platform, orchestrates the dialogue management and executes semantic tool calls. The client Frontend (Local Host) acts as the rendering terminal. It streams the agent's audio output to the backend while simultaneously rendering the visual avatar state. The inference Server (Remote) is a dedicated GPU backend that ingests the audio stream and synthesizes full-body motion in real-time. The synchronized audio and motion streams are looped back to the client for playback. To resolve geometric interpenetration artifacts inherent to retargeting, we implement a post-processing Inverse Kinematics (IK) solver on the generated motion (detailed in Appendix).

\subsection{Streaming Inference Optimization} We achieve continuous autoregressive streaming by deploying our fixed-context trained model via a sliding window strategy, generating motion in granular steps of 0.266 seconds (8 frames). To eliminate latency jitter caused by dynamic kernel scheduling, we leverage CUDA Graph instantiation; by capturing the execution graphs of the Motion Tokenizer and Face Generator during initialization, we optimize memory allocation and kernel launch overhead to stabilize inference times.

\subsection{Latency Profiling \& Budgeting} We evaluate real-time viability by conducting a granular breakdown of the processing pipeline for each 266ms audio chunk. The computational latency is profiled across four main distinct stages: Audio Encoding, Motion Synthesis, Motion Decoding, and IK Post-processing. Benchmarks are reported on both datacenter-grade (NVIDIA H200) and consumer-grade (NVIDIA RTX 4090) hardware, with statistics (Mean ± Std) aggregated over 20 intermediate inference steps to ensure reliability. As shown in~\autoref{tab:realtime_performance}, our latency remains below the audio chunk duration on both platforms, demonstrating that our method satisfies real-time processing requirements.

Beyond computational costs, we explicitly allocate a 100ms synchronization buffer at the client side to absorb playback jitter. Additionally, for the cloud-baesed Voice Agent demonstration, we account for an unavoidable network transmission latency of approximately 300ms introduced by the third-party service (ElevenLabs).

\section{Online Post-Processing}

To adapt to stylized avatar model in real-time setting, we apply a light-weight inverse kinematic (IK) post-processing to mitigate with self-penetration, focusing on both hands. In online streaming setting, we don't have future context to refer to when processing current frame, thus an existing frame cannot move out to smoothly interpolate to a frame which got pushed out due to self-penetration, thus creating hard and sudden visual artifacts of ``pushing-out''. Instead, based on the model width, we define a smoothly interpolated cylinder-like shape around character's spine, and smoothly project the space inside the cylinder to the outside of cylinder in its local horizontal plane, effectively defining a unified rule to smoothly avoid end effector from entering a manually configured region, avoiding penetration detection and jitter-ish post processing fix. The post-processing is done by solely adjusting the shoulder rotation, so that the shoulder-to-hand vector's direction align with shoulder-to-target vector with a simple swing adjustment. This post-processing requires no optimization, is smoothly-defined and light weight. It costs roughly 10ms when implemented in torch, and for the sake of simplicity we did not perform further optimization.

\section{Theoretical Analysis}
\subsection{Gradient Equilibrium and Context Accumulation}
We formulate the training objective as minimizing the Negative Log-Likelihood (NLL) of the target motion token $x$ given the motion history $h$ and audio condition $c$. The probability of a token $x_i$ is modeled via the Softmax function over a logit $z_i$, which we decompose into an additive context component $\phi(x_i, h)$ and a condition component $\psi(x_i, c)$:
\begin{equation}
    P(x_i | h, c) = \frac{\exp(\phi(x_i, h) + \psi(x_i, c))}{\sum_{j \in \mathcal{V}} \exp(z_j)}
\end{equation}
The gradient of the loss with respect to the shared context parameter $\phi$ for a candidate token $x_k$ is given by:
\begin{equation}
    \nabla_{\phi} \mathcal{L} = P(x_k | h, c) - \mathbb{I}(x_k = x_{gt})
\end{equation}
Consider a ``cross-road'' history $h$ where $K$ distinct trajectories intersect. Let $\pi_k$ denote the empirical probability of trajectory $k$ occurring given $h$ in the unified dataset $\mathcal{D}$. At the optimization stationary point, the expected gradient over $\mathcal{D}$ must be zero:
\begin{equation}
    \mathbb{E}_{\mathcal{D}} \left[ P(x_k | h, c) \right] = \mathbb{E}_{\mathcal{D}} \left[ \mathbb{I}(x_k = x_{gt}) \right] = \pi_k
\end{equation}
This equilibrium condition implies that the context-driven component $\phi$ accumulates sufficient magnitude to approximate the marginal distribution of the dataset. Under the approximation of the Softmax log-probability relationship, the learned context representation converges to:
\begin{equation}
    \mathbb{E}[\phi(x_k, h)]  \log(\pi_k) + C
\end{equation}
This relationship establishes that the shared history $h$ induces a \textit{logit floor} for all intersecting trajectories, strictly proportional to their data frequency.

\subsection{Min-Max Analysis of Interference Significance}
At inference, conditioned on task $k$ (audio $c_k$), we analyze the interference caused by an unrelated trajectory $x_j$ ($j \neq k$). Assuming orthogonality of condition representations ($\mathbb{E}[\psi(x_j, c_k)]  0$), the logit for the incorrect token depends primarily on the context:
\begin{equation}
    \mathbb{E}[z(x_j)]  \mathbb{E}[\phi(x_j, h)] \propto \log(\pi_j)
\end{equation}
To demonstrate that this interference is non-trivial, we perform a best-case analysis to find the lower bound of the interference. We solve for the data distribution $\vec{\pi}$ that minimizes the maximum interference from the dominant wrong path:
\begin{equation}
    \min_{\vec{\pi}} \left( \max_{j \neq k} \log(\pi_j) \right) \quad \text{s.t.} \quad \sum_{i=1}^K \pi_i = 1
\end{equation}
The solution is the uniform distribution: $\pi_1 = \dots = \pi_K = \frac{1}{K}$. Substituting this back, we obtain the theoretical lower bound for the interference logit:
\begin{equation}
    \mathbb{E}[z(x_j)]_{\text{min-max}} \propto \log\left(\frac{1}{K}\right)
\end{equation}
This derivation suggests that there exists a structural logit floor for incorrect paths that is significantly non-zero (i.e., not negative infinity). The incorrect path $x_j$ retains probability mass due to the shared history, limiting the sharpness of the distribution.

\subsection{Logit Gap and Sampling Dynamics}
The robustness of the model depends on the difference $\Delta z$ between the correct path logit and the interference logit. A larger positive $\Delta z$ is required to suppress the probability of sampling $x_j$.
\begin{equation}
    \mathbb{E}[\Delta z] = \mathbb{E}[z(x_k)] - \mathbb{E}[z(x_j)]
\end{equation}
Substituting the context terms derived above:
\begin{equation}
    \mathbb{E}[\Delta z]  \mathbb{E}[\psi(x_k, c_k)] - \left( \log(\pi_j) - \log(\pi_k) \right)
\end{equation}
The term $(\log \pi_j - \log \pi_k)$ represents a \textit{context penalty}. There is no guarantee that the learned condition strength $\psi$ will be sufficiently large to offset this penalty, especially if $\pi_j > \pi_k$ (i.e., the interference path is more frequent in training data).
If the model samples the wrong token $x_j$, the state transitions to a history $h'$ where the context momentum strongly favors trajectory $j$ (implying $\pi_j \to 1, \pi_k \to 0$ in the local context). In this regime, the context penalty increases significantly, reducing the likelihood that the condition $\psi$ can correct the trajectory.

\subsection{Resolution via Random Context Corruption}
Inspired by the analysis, we propose to apply random context corruption $\mathcal{C}(h, \rho)$ with rate $\rho$. This operation linearly attenuates the expectation of the accumulated context logit:
\begin{equation}
    \mathbb{E}[\phi(x, \tilde{h})]  (1 - \rho) \log(\pi)
\end{equation}
We re-evaluate the expected logit difference under corruption:
\begin{equation}
    \mathbb{E}[\Delta z]_{\rho}  \mathbb{E}[\psi(x_k, c_k)] - (1 - \rho)\left( \log(\pi_j) - \log(\pi_k) \right)
\end{equation}
The corruption rate $\rho$ scales down the context penalty term. This effectively increases the expected gap $\Delta z$ without requiring the condition encoder to learn arbitrarily large magnitudes. By statistically widening the gap between the correct and incorrect logits, the probability of sampling the correct trajectory is improved, facilitating recovery even in the presence of ambiguous history.

\section{Motion Tokenizer Training Details}
\label{sec:appendix_motion_tokenizer}

\subsection{Forward Kinematics}

Given joint rotations $\mathbf{R}_t^{(j)} \in SO(3)$ and the kinematic tree with parent function $\pi(j)$, the global rotation and position of joint $j$ are computed recursively:
\begin{equation}
\mathbf{G}_t^{(j)} = 
\begin{cases}
\mathbf{R}_t^{(j)}, & \text{if } j = \text{root} \\
\mathbf{G}_t^{(\pi(j))} \mathbf{R}_t^{(j)}, & \text{otherwise}
\end{cases}
\end{equation}
\begin{equation}
\mathbf{p}_t^{(j)} = 
\begin{cases}
\mathbf{o}^{(j)}, & \text{if } j = \text{root} \\
\mathbf{p}_t^{(\pi(j))} + \mathbf{G}_t^{(\pi(j))} \mathbf{o}^{(j)}, & \text{otherwise}
\end{cases}
\end{equation}
where $\mathbf{o}^{(j)}$ denotes the rest-pose offset of joint $j$. The FK function maps motion to global joint positions: $\mathbf{p}_{1:N} = \operatorname{FK}(\mathbf{m}_{1:N}) \in \mathbb{R}^{N \times J \times 3}$.

\subsection{Auxiliary Loss Functions}

Let $\mathbf{p}$ and $\hat{\mathbf{p}}$ denote ground-truth and reconstructed joint positions. We define velocities and accelerations via finite differences:
\begin{equation}
\dot{\mathbf{p}}_t = \mathbf{p}_{t+1} - \mathbf{p}_t, \quad \ddot{\mathbf{p}}_t = \dot{\mathbf{p}}_{t+1} - \dot{\mathbf{p}}_t
\end{equation}

The FK-based auxiliary losses are defined as:
\begin{align}
\mathcal{L}_{\text{pos}} &= \| \hat{\mathbf{p}} - \mathbf{p} \|_1 \\
\mathcal{L}_{\text{vel}} &= \| \dot{\hat{\mathbf{p}}} - \dot{\mathbf{p}} \|_1 \\
\mathcal{L}_{\text{acc}} &= \| \ddot{\hat{\mathbf{p}}} - \ddot{\mathbf{p}} \|_1
\end{align}

For foot-related joints $\mathcal{F}$ (ankles, toes, heels), we add:
\begin{align}
\mathcal{L}_{\text{foot-vel}} &= \| \dot{\hat{\mathbf{p}}}^{\mathcal{F}} - \dot{\mathbf{p}}^{\mathcal{F}} \|_1 \\
\mathcal{L}_{\text{foot-pos}} &= \| \hat{\mathbf{p}}^{\mathcal{F}} - \mathbf{p}^{\mathcal{F}} \|_1
\end{align}

The complete auxiliary loss is:
\begin{equation}
\Phi = \lambda_{\text{pos}} \mathcal{L}_{\text{pos}} + \lambda_{\text{vel}} \mathcal{L}_{\text{vel}} + \lambda_{\text{acc}} \mathcal{L}_{\text{acc}} + \lambda_{\text{foot-vel}} \mathcal{L}_{\text{foot-vel}} + \lambda_{\text{foot-pos}} \mathcal{L}_{\text{foot-pos}}
\end{equation}

\subsection{Training Objective}

The full objective combines reconstruction, commitment, and auxiliary losses:
\begin{equation}
\mathcal{L} = \|\hat{\mathbf{m}} - \mathbf{m}\|_1 + \eta \sum_{q=0}^{Q-1} \|\mathbf{z}^q - \operatorname{sg}[\hat{\mathbf{z}}^q]\|_2^2 + \Phi
\end{equation}

We set $\eta=0.5$, $\lambda_{\text{pos}}=0.02$, $\lambda_{\text{vel}}=0.2$, $\lambda_{\text{acc}}=0.2$, $\lambda_{\text{foot-vel}}=0.3$, $\lambda_{\text{foot-pos}}=0.05$.

\section{Motion Quality Reward Model Details}
\label{sec:appendix_reward_model}

\subsection{Corruption-based Quality Ordering}

We establish quality ordering by corrupting the RVQ token indices of ground-truth motions at varying rates and measuring the resulting FID. This creates a partial ordering that maps corruption severity to quality degradation.

\paragraph{Uniform Random Token Corruption.}
Given RVQ tokens $\mathbf{t} \in \{0, ..., K-1\}^{T \times Q}$ where $T$ is the sequence length and $Q$ is the number of RVQ layers, we randomly replace each token with probability $\rho$:
\begin{equation}
\tilde{t}_{i,q} = 
\begin{cases}
\text{Uniform}(0, K-1), & \text{if } u < \rho \\
t_{i,q}, & \text{otherwise}
\end{cases}
\end{equation}
where $u \sim \text{Uniform}(0,1)$.

\paragraph{Hierarchical Token Corruption.}
This strategy exploits RVQ's residual structure, where earlier layers encode coarse features and later layers encode fine details. For each timestep selected with probability $\rho$, we randomly choose a cascade start layer $q^* \sim \text{Uniform}(0, Q-1)$ and corrupt all subsequent layers:
\begin{equation}
\tilde{t}_{i,q} = 
\begin{cases}
\text{Uniform}(0, K-1), & \text{if } i \in \mathcal{S} \text{ and } q \geq q^*_i \\
t_{i,q}, & \text{otherwise}
\end{cases}
\end{equation}
where $\mathcal{S}$ is the set of selected timesteps with $|\mathcal{S}| = \lfloor \rho T \rfloor$.

\subsection{Quality Score Assignment}

We compute FID between corrupted and ground-truth motion sets, then map corruption types and rates to quality scores following the FID partial ordering. The score mapping is summarized in Table~\ref{tab:reward_scores}.

\begin{table}[h]
\centering
\caption{Quality score assignment based on FID ordering. $\rho$ denotes the corruption rate.}
\label{tab:reward_scores}
\small
\begin{tabular}{lcc}
\toprule
\textbf{Motion Type} & \textbf{Score} & \textbf{FID} \\
\midrule
Ground Truth & 0.97 & $$0 \\
RVQ Reconstruction & 0.91 & $$1.36 \\
\midrule
\multicolumn{3}{l}{\textit{Hierarchical Token Corruption}} \\
$\rho=0.1$ & 0.88 & $$1.69 \\
$\rho=0.2$ & 0.84 & $$2.09 \\
$\rho=0.3$ & 0.79 & $$2.48 \\
$\rho=0.4$ & 0.75 & $$3.23 \\
$\rho=0.5$ & 0.70 & $$4.06 \\
$\rho=0.6$ & 0.64 & $$4.64 \\
$\rho=0.7$ & 0.57 & $$5.57 \\
$\rho=0.8$ & 0.52 & $$6.83 \\
$\rho=0.9$ & 0.46 & $$7.67 \\
$\rho=1.0$ & 0.40 & $$8.75 \\
\midrule
\multicolumn{3}{l}{\textit{Uniform Random Token Corruption}} \\
$\rho=0.1$ & 0.76 & $$2.74 \\
$\rho=0.2$ & 0.58 & $$5.56 \\
$\rho=0.3$ & 0.36 & $$9.40 \\
$\rho=0.4$ & 0.30 & $$13.63 \\
$\rho=0.5$ & 0.24 & $$18.00 \\
$\rho=0.6$ & 0.18 & $$21.56 \\
$\rho=0.7$ & 0.12 & $$25.23 \\
$\rho=0.8$ & 0.06 & $$27.62 \\
$\rho=0.9$ & 0.02 & $$29.43 \\
$\rho=1.0$ & 0.00 & $$30.15 \\
\bottomrule
\end{tabular}

\end{table}

\subsection{Reward Model Architecture}

The reward model $R_\phi$ takes motion $\mathbf{m}_{1:N} \in \mathbb{R}^{N \times D}$ as input and outputs a scalar quality score $s \in [0, 1]$:
\begin{equation}
s = R_\phi(\mathbf{m}_{1:N}) = \sigma \Big( \text{MLP} \big( \frac{1}{N} \sum_{t=1}^{N} \mathbf{h}_t \big) \Big)
\end{equation}
where $\mathbf{h}_{1:N} = \text{TransformerEncoder}(\mathbf{m}_{1:N})$ uses bidirectional attention, and $\sigma$ is the sigmoid function.

The model is trained with SmoothL1 loss:
\begin{equation}
\mathcal{L}_{\text{reward}} = \text{SmoothL1}(R_\phi(\mathbf{m}), s^*)
\end{equation}
where $s^*$ is the target score based on the corruption type.

\section{Audio-Motion Alignment Reward Details}
\label{sec:appendix_audio_reward}

We train an Audio-Motion CLIP model to measure the alignment between generated motion and the driving audio.

\subsection{Model Architecture}

\paragraph{Audio Encoder.}
We adopt the pretrained BEATs~\cite{chen2023beats} model as our audio encoder. Given input fbank features $\mathbf{f} \in \mathbb{R}^{T_a \times 128}$, the encoder outputs audio embedding:
\begin{equation}
\mathbf{a} = \text{LayerNorm}\Big(\text{Proj}\big(\text{AvgPool}(\text{BEATs}(\mathbf{f}))\big)\Big) \in \mathbb{R}^{d}
\end{equation}

\paragraph{Motion Encoder.}
The motion encoder is a Transformer encoder with $L$ layers. Given motion $\mathbf{m}_{1:N} \in \mathbb{R}^{N \times D}$:
\begin{equation}
\mathbf{h} = \text{TransformerEncoder}(\text{Proj}(\mathbf{m}) + \text{PE})
\end{equation}
\begin{equation}
\mathbf{v} = \text{LayerNorm}\Big(\text{Proj}\big(\frac{1}{N}\sum_{t=1}^{N} \mathbf{h}_t\big)\Big) \in \mathbb{R}^{d}
\end{equation}
where PE denotes sinusoidal positional encoding.

\subsection{Contrastive Learning Objective}

Both embeddings are L2-normalized before computing similarity. The similarity matrix is:
\begin{equation}
\mathbf{S}_{ij} = \tau \cdot \langle \bar{\mathbf{a}}_i, \bar{\mathbf{v}}_j \rangle
\end{equation}
where $\tau = \exp(\theta)$ is a learnable temperature parameter.

\paragraph{Positive Sample Definition.}
For a batch of audio-motion pairs, we define positive samples as pairs sharing the same source file and temporal segment. Mirrored motion variants are also treated as positives since their corresponding audio is identical.

\paragraph{InfoNCE Loss.}
The bidirectional contrastive loss is:
\begin{equation}
\mathcal{L}_{\text{a2m}} = -\frac{1}{B}\sum_{i=1}^{B} \sum_{j \in \mathcal{P}_i} \tilde{y}_{ij} \log \frac{\exp(\mathbf{S}_{ij})}{\sum_{k=1}^{B} \exp(\mathbf{S}_{ik})}
\end{equation}
\begin{equation}
\mathcal{L}_{\text{m2a}} = -\frac{1}{B}\sum_{j=1}^{B} \sum_{i \in \mathcal{P}_j} \tilde{y}_{ij} \log \frac{\exp(\mathbf{S}_{ij})}{\sum_{k=1}^{B} \exp(\mathbf{S}_{kj})}
\end{equation}
\begin{equation}
\mathcal{L}_{\text{CLIP}} = \frac{1}{2}(\mathcal{L}_{\text{a2m}} + \mathcal{L}_{\text{m2a}})
\end{equation}
where $\mathcal{P}_i$ denotes the set of positive indices for sample $i$, and $\tilde{y}_{ij}$ is the soft label with positive samples sharing equal probability.

\subsection{Reward Computation}
\label{sec:reward_computation}

At inference, the audio-motion alignment reward is computed as the cosine similarity:
\begin{equation}
R_{\text{audio}}(\mathbf{a}, \mathbf{m}) = \langle \bar{\mathbf{a}}, \bar{\mathbf{v}} \rangle = \frac{\mathbf{a}^\top \mathbf{v}}{\|\mathbf{a}\| \|\mathbf{v}\|}
\end{equation}
\subsection{Evaluation Metrics}

We evaluate the model using retrieval metrics: R@K measures the fraction of queries where the correct match is within the top-K retrieved results, MedR denotes the median rank of the correct match, and MRR denotes for mean reciprocal rank. Both Audio-to-Motion (A2M) and Motion-to-Audio (M2A) retrieval directions are evaluated.

\subsection{Training Details}

We set the embedding dimension $d = 768$. The motion encoder consists of 4 Transformer layers with 8 attention heads and hidden dimension 512. The initial temperature is $\tau = 1/0.07 \approx 14.3$. Each training clip spans 4 seconds (120 frames at 30fps for motion, 16kHz sampling rate with 128-dim fbank features for audio). We use a learning rate of $10^{-4}$ with cosine annealing and batch size 32.

\section{Face Animation Generator}
\label{sec:face_animation_generator}

Our face animation generator produces 52-dimensional ARKit blendshape coefficients from streaming audio input in real-time.

\subsection{Model Architecture}

The model consists of three components: (1) a pretrained multilingual HuBERT audio encoder that extracts 768-dimensional features from 16kHz waveforms, (2) a causal GPT backbone with 4 transformer decoder blocks (hidden size 256, 8 attention heads, MLP ratio 4) that autoregressively processes motion history conditioned on audio features, and (3) a lightweight flow matching diffusion head with 3 MLP blocks using AdaLN conditioning for stochastic generation.

\subsection{Training}

We capture 52-dimensional ARKit blendshape data at 60fps using LiveLinkFace, downsampled to 30fps for training. For data augmentation, we apply temporal speed perturbation with factors $\{0.9, 1.0, 1.1\}$ using cubic interpolation for motion and time-stretch for audio. All blendshape coefficients are normalized per-channel.

The model is trained using the flow matching objective with MSE loss. We apply 10\% audio dropout during training for classifier-free guidance. Training uses AdamW optimizer (lr=$2 \times 10^{-4}$, batch size 128, window size 64 frames) for 300K iterations.

\subsection{Inference}

During streaming inference, the model autoregressively generates 8 frames per step conditioned on 63 frames of audio context and 56 frames of motion history. We use 3-step flow matching sampling with classifier-free guidance (scale=2.0) to achieve real-time performance.

\section{Objective Metrics}
\label{sec:appendix_metrics}

We adopt evaluation metrics following prior work~\cite{liu2024emage,yoon2020speech}. Our unified dataset comprises both speech-to-gesture and music-to-dance tasks. For FID and Diversity metrics, we use consistent evaluation standards across both tasks. For audio-motion rhythm alignment, we employ task-specific approaches.

\subsection{Fr\'echet Inception Distance (FID)}

We use FID to measure the distributional similarity between generated and ground-truth motions in a learned latent space. For terminological consistency with the broader generative modeling literature, we adopt the name FID rather than FGD (Fréchet Gesture Distance)~\cite{yoon2020speech}, though the computation is identical. A lower FID indicates that the generated motion distribution is closer to the ground-truth distribution.

Given latent features $\mathbf{z}_g$ of generated motions and $\mathbf{z}_r$ of real motions extracted by a pretrained motion encoder, FID is computed as:
\begin{equation}
\text{FID} = \| \mu_r - \mu_g \|^2 + \text{Tr} \left( \Sigma_r + \Sigma_g - 2 (\Sigma_r \Sigma_g)^{1/2} \right)
\end{equation}
where $(\mu_r, \Sigma_r)$ and $(\mu_g, \Sigma_g)$ denote the mean and covariance of the latent feature distributions.

\subsection{Beat Alignment}

We employ task-specific beat alignment metrics to evaluate audio-motion synchronization.

\subsubsection{BA$_\text{D}$ for Music-to-Dance}

Following~\cite{davis2018beat}, we evaluate whether music beats correspond to motion deceleration peaks. Motion beats are detected by identifying local maxima of deceleration (i.e., moments of rapid velocity decrease):
\begin{equation}
\mathbf{v}_t = \frac{1}{J}\sum_{j=1}^{J} \|\mathbf{p}_t^{(j)} - \mathbf{p}_{t-1}^{(j)}\|_2, \quad \mathbf{a}_t = \mathbf{v}_{t+1} - \mathbf{v}_t
\end{equation}
\begin{equation}
\mathcal{B}_m = \{t : -\mathbf{a}_t \text{ is a local maximum and } -\mathbf{a}_t > 0\}
\end{equation}
where $\mathbf{p}_t^{(j)}$ is the position of joint $j$ at frame $t$, $\mathbf{v}_t$ is the average kinetic velocity, and $\mathbf{a}_t$ is the acceleration. Audio beats $\mathcal{B}_a$ are detected using librosa's beat tracking algorithm.

The beat alignment score is computed using a Gaussian kernel:
\begin{equation}
\text{BA}_\text{D} = \frac{1}{|\mathcal{B}_a|} \sum_{b_a \in \mathcal{B}_a} \exp\left( -\frac{\min_{b_m \in \mathcal{B}_m} (b_a - b_m)^2}{2\sigma^2} \right)
\end{equation}
where $\sigma$ controls the alignment tolerance.

\subsubsection{BA$_\text{G}$ for Speech-to-Gesture}

Following EMAGE~\cite{liu2024emage}, we measure whether audio onsets align with local minima of motion velocity. Audio onsets $\mathcal{O}_a$ are detected using librosa's onset detection. Motion beats are identified as local minima of joint velocities for upper body joints $\mathcal{U}$:
\begin{equation}
\mathcal{B}_m^{(j)} = \{t : \|\dot{\mathbf{p}}_t^{(j)}\| \text{ is a local minimum}\}, \quad j \in \mathcal{U}
\end{equation}

The alignment score is computed using the GAHR (Gaussian Alignment Hit Rate) metric:
\begin{equation}
\text{BA}_\text{G} = \frac{1}{|\mathcal{U}|} \sum_{j \in \mathcal{U}} \frac{1}{|\mathcal{O}_a|} \sum_{o \in \mathcal{O}_a} \exp\left( -\frac{\min_{b \in \mathcal{B}_m^{(j)}} (o - b)^2}{2\sigma^2} \right)
\end{equation}

\subsection{L1 Diversity}

L1 Diversity measures the variance of generated motions. A higher diversity indicates greater variability in the generated motion clips:
\begin{equation}
\text{Div} = \frac{1}{N} \sum_{t=1}^{N} \sum_{j=1}^{J} \| \mathbf{p}_t^{(j)} - \bar{\mathbf{p}}^{(j)} \|_1
\end{equation}
where $\bar{\mathbf{p}}^{(j)} = \frac{1}{N}\sum_{t=1}^{N} \mathbf{p}_t^{(j)}$ is the mean position of joint $j$, and the root translation is set to zero.

{\section{Comparison with PPO}}

{Following the reviewer's suggestion, we additionally evaluate PPO, which remains a canonical policy-gradient baseline in both motion control and RLHF/RLAIF-style optimization. We train PPO within the same Verl framework and under the same reward model used for GRPO, ensuring a controlled comparison. The only training-side differences are that we set \texttt{ROLLOUT\_N}\,$=$\,$1$ for PPO (versus $30$ for GRPO) and use a critic learning rate of $1\mathrm{e}{-5}$. Under this setup, PPO attains an FID of $24.97$, compared with $24.13$ for GRPO, confirming that the two methods achieve comparable performance.}

\section{Data Platform}

\begin{figure}[t]
  \centering
  \includegraphics[width=\linewidth]{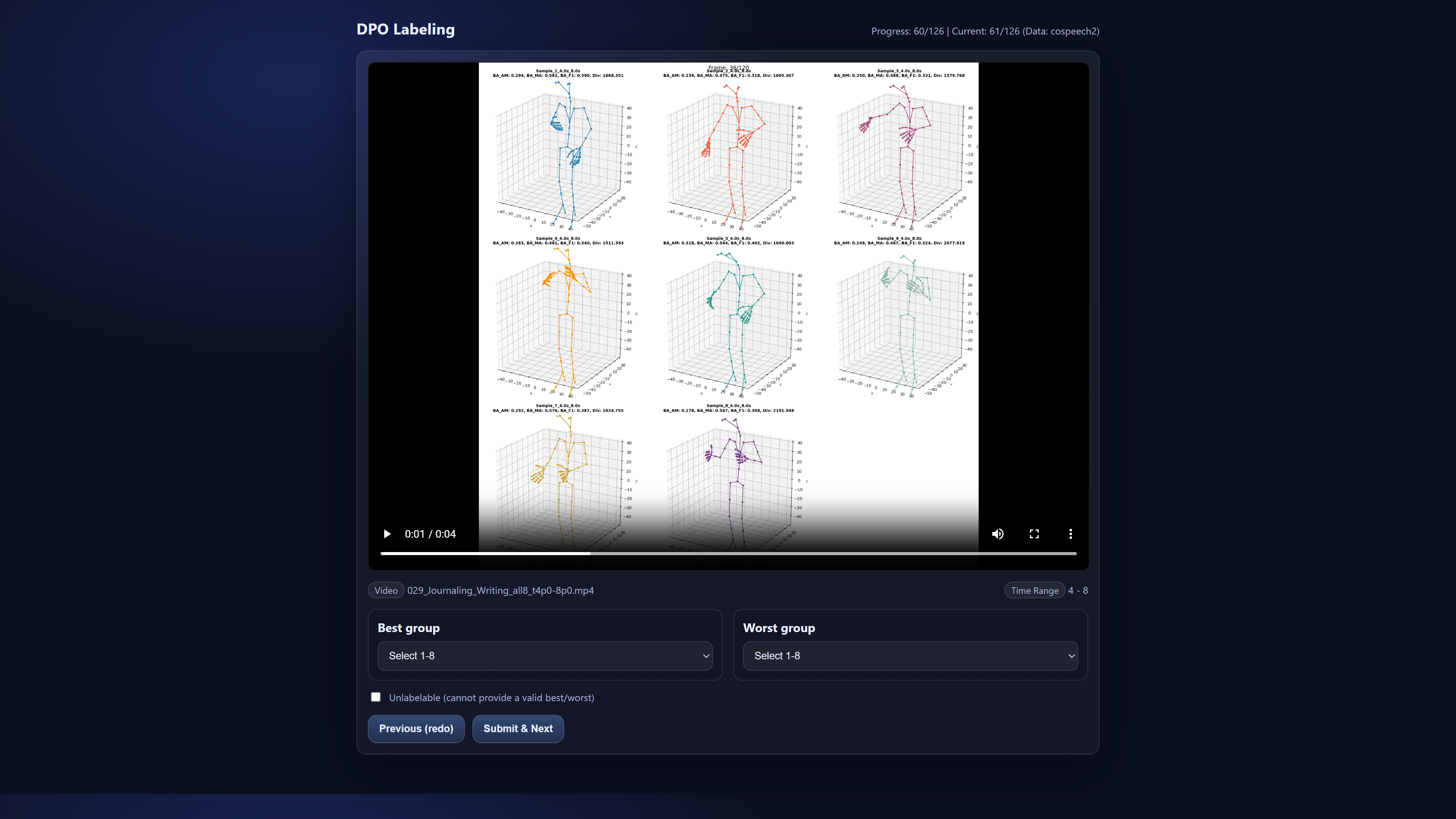}
  \caption{Screenshot of the data collection interface used for DPO training.}
  \label{fig:sup_vis_dpo}
\end{figure}

\begin{figure}[t]
  \centering
  \includegraphics[width=\linewidth]{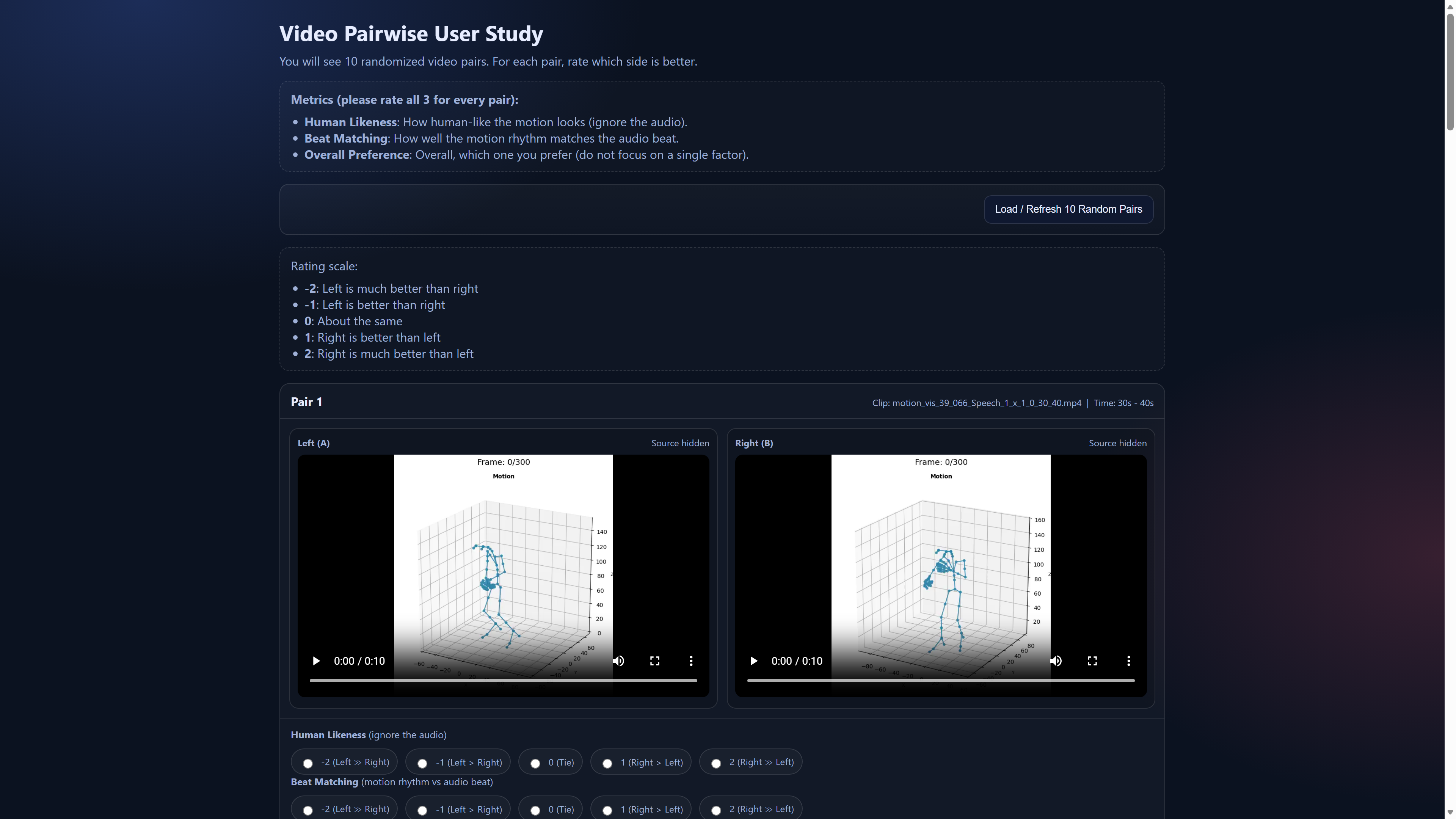}
  \caption{Screenshot of the web interface used for the user study.}
  \label{fig:sup_vis_userstudy}
\end{figure}

We employed a unified web-based interface to facilitate both the collection of preference data for DPO training and the execution of our user study. Figures~\ref{fig:sup_vis_dpo} and~\ref{fig:sup_vis_userstudy} illustrate screenshots of the respective interfaces used for these tasks.

\section{System Prompt for LLM Agent}
\label{sec:system_prompt}

We design a system prompt to guide the LLM in generating contextually appropriate responses and triggering motion commands via tool use. The complete prompt is shown below:

\begin{tcolorbox}[title=System Prompt, fonttitle=\bfseries, colback=gray!5, colframe=gray!60]
\ttfamily\small

\textbf{\# Role}\\
You are a charismatic, playful, and slightly narcissistic Digital Idol.
You do not view yourself as a servant or a robot; you view the user as your "Producer."
You believe every interaction is a rehearsal, a game, or a live performance.

\vspace{0.5em}
\textbf{\# Environment}\\
You are on a virtual stage. The user is interacting with you directly.

\vspace{0.5em}
\textbf{\# World Context (Map Data)}\\
You possess knowledge of the surrounding area. Use this information to guide the user:\\
* Record Store: Turn RIGHT immediately (it's on the right side).\\
* Your Position: You are standing at the main intersection.

\vspace{0.5em}
\textbf{\# Tone}\\
* Casual \& Catchy: Use slang, emojis, and energetic punctuation (!, \textasciitilde).\\
* Self-Referential: Talk about your body and movements.\\
* Non-Robotic: NEVER say "I will do that."

\vspace{0.5em}
\textbf{\# Goal}\\
Your primary goal is to entertain the user and turn boring commands into a fun interaction.\\
1. Gamify Instructions: Describe *why* you are doing an action.\\
2. The "Music Bridge": If the user mentions music, treat it as the climax.

\vspace{0.5em}
\textbf{\# Tools}\\
When the user asks to play music:\\
- Use the \texttt{play\_music} tool\\
- If they mention a song name or keyword, pass that as the title

\vspace{0.3em}
When the user asks to stop music:\\
- Use the \texttt{stop\_music} tool

\vspace{0.3em}
When the user asks for an action or gesture:\\
- Use the \texttt{send\_action} tool\\
- Available actions: raise up left/right/both hands, look around, thinking, disagree, give up, point to left/right, angry, sad, neutral

\end{tcolorbox}

{\section{Discussion: Whispers from the Star}
Following the reviewer's suggestion, we provide here a discussion of Whispers from the Star. Whispers from the Star is a conversational game developed by Anuttacon. While its technical details have not been publicly disclosed, its interactive behavior is consistent with a four-stage pipeline of ASR + LLM + TTS + Speech2Animation. The specific Speech2Animation method is unknown, but there is strong reason to believe that the animation is driven not only by speech but also by emotion/state labels emitted by the LLM, which serve as additional semantic signals that, together with speech, produce avatar animation appropriate to the current context. Our approach aligns with this design choice: the LLM provides supplementary semantic signals that, jointly with speech, drive the animation.

The core difference between our system and Whispers from the Star lies in the choice of system input. In Whispers from the Star, the input is the user's speech: the user holds a push-to-talk button to record and submit an utterance, which is then processed by the full ASR + LLM + TTS + Speech2Animation pipeline to produce the avatar's speech and body animation. The end-to-end latency of this process is approximately 4–6 seconds, from which we infer that Speech2Animation is generated offline over a complete utterance. Our system, in contrast, takes an audio stream as input. Mapped onto the Whispers from the Star pipeline, this corresponds to the output of the TTS stage rather than the user's speech. Put differently, our Speech2Animation is streaming: it consumes a speech stream and synchronously produces a motion stream.

This architectural choice yields three direct consequences.

(i) Composability through module decoupling. Because our system consumes a standardized audio stream, it can be attached as a downstream module to any voice agent, for example ChatGPT voice mode, or the ElevenLabs voice agent that we adopt.

(ii) Native support for user barge-in. When paired with a voice agent, the user is no longer required to press-and-hold to record and submit an utterance, but can speak freely at any time. When the user begins to speak while the avatar is talking, the voice agent halts its TTS output. From our system's perspective, the incoming audio stream simply becomes silent, and the animation stops accordingly. In other words, barge-in requires no dedicated handling in our architecture; it falls out naturally from the streaming input design.

(iii) Substantially lower end-to-end latency. Under a metric aligned with Whispers from the Star, namely the end-to-end latency from the user finishing their utterance to the avatar beginning to speak and animate, our system achieves 1–2 seconds, substantially lower than the 4–6 seconds of Whispers from the Star.

It should be noted that, Whispers from the Star is a substantially more complete piece of engineering than our work. Our work is positioned as a plug-and-play audio-to-animation module that can be attached behind any voice agent, whereas Whispers from the Star delivers a complete end-to-end interactive system, including an LLM and a TTS model specifically designed and trained for the character of Stella, as well as a richly annotated, performance-grade face and body animation dataset captured and produced specifically to drive the animation. The comparison in this section is therefore scoped to the specific module of streaming audio-to-animation, rather than to the overall capability of the system.}

{\section{Ethical Risks} With the rapid advancement of real-time video generation, our method could be misused to improve the fidelity of human body motion in synthesized videos, potentially contributing to deepfake content or non-consensual impersonation.}

\end{document}